\documentclass[10pt,twocolumn,letterpaper]{article}

\usepackage[pagenumbers]{cvpr}

\usepackage{graphicx}
\usepackage{amsmath}
\usepackage{amssymb}
\usepackage{bm} 
\usepackage{upgreek} 
\usepackage{booktabs}
\usepackage{float}
\usepackage{caption}
\usepackage{placeins}
\usepackage{enumitem}

\usepackage[pagebackref,breaklinks,colorlinks]{hyperref}

\usepackage[capitalize]{cleveref}
\crefname{section}{Sec.}{Secs.}
\Crefname{section}{Section}{Sections}
\Crefname{table}{Table}{Tables}
\crefname{table}{Tab.}{Tabs.}
\renewcommand{\figureautorefname}[1]{Fig.\,}
\renewcommand{\equationautorefname}[1]{Eq.\,}

\DeclareMathOperator*{\argmax}{arg\,max}

\newcommand*\rot{\rotatebox{90}}

\renewcommand\vec{\mathbf}
\newcommand\wc{\vec{w}_{\mathrm{c}}}
\newcommand\fc{f_{\mathrm{c}}}
\newcommand\zvec{\vec{z}} 
\newcommand\wvec{\vec{w}}
\newcommand\muvec{\bm{\upmu}} 
\DeclareMathOperator{\Tr}{Tr}

\def\cvprPaperID{*****} 
\def\confName{CVPR}
\def\confYear{2022}

\newsavebox\commabox
\sbox{\commabox}{,}
\newcommand\fnover{\kern-\wd\commabox} 

\begin{document}

\title{Art Creation with Multi-Conditional StyleGANs}
\author{Konstantin Dobler,\fnover$^1$ Florian Hübscher,\fnover$^1$ Jan Westphal,\fnover$^1$\\Alejandro Sierra-Múnera,\fnover$^1$ Gerard de Melo,\fnover$^1$ Ralf Krestel$^2$\\~\\
1:~ Hasso Plattner Institute, University of Potsdam\\
2:~ ZBW -- Leibniz Information Centre for Economics and Kiel University\\
}
\maketitle

\begin{abstract}
Creating meaningful art is often viewed as a uniquely human endeavor.
A human artist needs a combination of unique skills, understanding, and genuine intention to create artworks that evoke deep feelings and emotions.
In this paper, we introduce a multi-conditional Generative Adversarial Network (GAN) approach trained on large amounts of human paintings to synthesize realistic-looking paintings that emulate human art.
Our approach is based on the StyleGAN neural network architecture, but incorporates a custom multi-conditional control mechanism that provides fine-granular control over characteristics of the generated paintings, e.g., with regard to the perceived emotion evoked in a spectator. 
For better control, we introduce the conditional truncation trick, which adapts the standard truncation trick for the conditional setting and diverse datasets.
Finally, we develop a diverse set of evaluation techniques tailored to multi-conditional generation.
\end{abstract}

\section{Introduction}
\label{sec:intro}

\begin{figure}[ht]
  \centering
  \begin{subfigure}{0.48\linewidth}
    \includegraphics[width=1\linewidth]{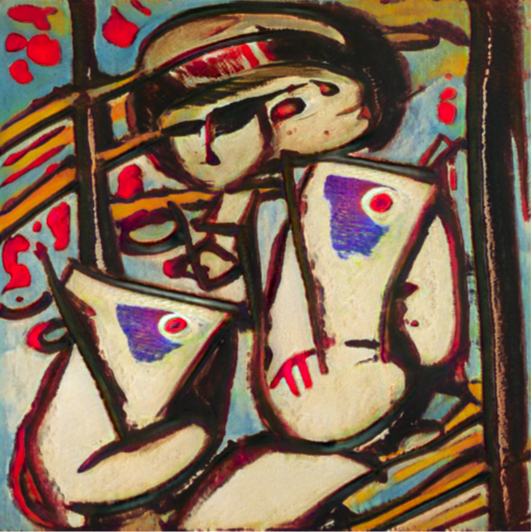}
  \end{subfigure}
  \hfill
  \begin{subfigure}{0.48\linewidth}
    \includegraphics[width=1\linewidth]{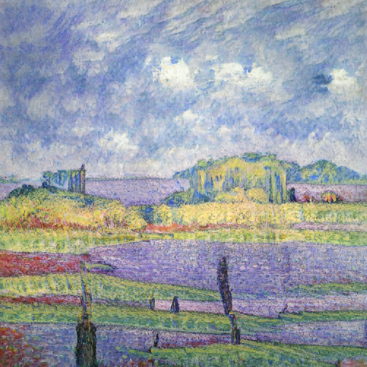}
  \end{subfigure}
  \caption{Example artworks produced by our StyleGAN models trained on the EnrichedArtEmis dataset (described in Section~\ref{sec:dataset}).}
  \label{fig:example-images-first-page}
\end{figure}

Creativity is an essential human trait and the creation of art in particular is often deemed a uniquely human endeavor. To create meaningful works of art, a human artist requires a combination of specific skills, understanding, and genuine intention. Such artworks may then evoke deep feelings and emotions. 
In light of this, there is a long history of endeavors to emulate this  computationally, starting with early algorithmic approaches to art generation in the 1960s.
Only recently, however, with the success of deep neural networks in many fields of artificial intelligence, has an automatic generation of images reached a new level. With new neural architectures and  massive compute, recent methods have been able to synthesize photo-realistic faces~\cite{karras-stylegan2-ada}. 

In this paper, we investigate models that attempt to create works of art resembling human paintings. We propose techniques that allow us to specify a series of conditions such that the model seeks to create images with particular traits, e.g., particular styles, motifs, evoked emotions, etc.

We adopt the well-known Generative Adversarial Network (GAN) framework~\cite{goodfellow2014generative}, in particular the StyleGAN2-ADA architecture~\cite{karras-stylegan2-ada}.
The greatest limitations until recently have been the low resolution of generated images as well as the substantial amounts of required training data.
This is exacerbated when we wish to be able to specify multiple conditions, as there are even fewer training images available for each combination of conditions.
We train our GAN using an enriched version of the ArtEmis dataset by Achlioptas~\etal~\cite{achlioptas2021artemis} and investigate the effect of multi-conditional labels.
Two example images produced by our models can be seen in \autoref{fig:example-images-first-page}.
Our contributions include:
\begin{itemize}[noitemsep,nolistsep]
    \item We explore the use of StyleGAN to emulate human art, focusing in particular on the less explored conditional capabilities,
    to control traits such as art style, genre, and content. We further investigate evaluation techniques for multi-conditional GANs.
    \item We introduce the concept of conditional center of mass in the StyleGAN architecture and explore its various applications. In particular, we propose a conditional variant of the \textit{truncation trick}~\cite{brock2018largescalegan} for the StyleGAN architecture that preserves the conditioning of samples.
    \item We formulate the need for \textit{wildcard generation} in multi-conditional GANs, and propose a method to enable wildcard generation by replacing parts of a multi-condition-vector during training.
\end{itemize}

\section{Related Work}
\label{sec:related}

Generative adversarial networks (GANs) \cite{goodfellow2014generative} are among the most well-known family of network architectures. 
The objective of the architecture is to approximate a target distribution, which,
in our setting, implies that the GAN seeks to produce images similar to those in the target distribution given by a set of training images.

GANs achieve this through the interaction of two neural networks, the generator $G$ and the discriminator $D$.
The generator produces \emph{fake} data, while the discriminator attempts to tell apart such generated data from genuine original training images. During training, as the two networks are tightly coupled, they both improve over time until $G$ is ideally able to approximate the target distribution to a degree that makes it hard for $D$ to distinguish between genuine original data and fake generated data.

Building on this idea, Radford~\etal combined convolutional networks with GANs to produce images of higher quality~\cite{radford2016unsupervised}.
Arjovsky~\etal proposed the Wasserstein distance, a new loss function under which the training of a Wasserstein GAN (WGAN) improves in stability and the generated images increase in quality~\cite{arjovsky2017wasserstein}.
Karras~\etal introduced progressive growing of GANs, a new training technique that allowed the generation of images of unprecedented resolution~\cite{karras2018progressive}.
They achieved this by progressively increasing the number of layers in the model to reduce the training time and to produce finer details.

With StyleGAN, that is based on style transfer, Karras~\etal presented a new GAN architecture~\cite{karras2019stylebased}
that improved the state-of-the-art image quality and provides control over both high-level attributes as well as finer details. 
Karras~\etal further improved the StyleGAN architecture with StyleGAN2, which removes characteristic artifacts from generated images~\cite{karras-stylegan2}.
Later on, they additionally introduced an adaptive augmentation algorithm (ADA) to StyleGAN2 in order to reduce the amount of data needed during training~\cite{karras-stylegan2-ada}.
Due to its high image quality and the increasing research interest around it, we base our work on the StyleGAN2-ADA model.

\paragraph{Conditional GANs.}
In recent years, different architectures have been proposed to incorporate conditions into the GAN architecture. 
The first conditional GAN (cGAN) was proposed by Mirza and Osindero, where the condition information is one-hot (or otherwise) encoded into a vector~\cite{mirza2014conditional}.
This encoding is concatenated with the other inputs before being fed into the generator and discriminator.
Another approach uses an auxiliary classification head in the discriminator~\cite{odena2017conditional}. 
The cross-entropy between the predicted and actual conditions is added to the GAN loss formulation to guide the generator towards conditional generation. 

Current state-of-the-art architectures employ a projection-based discriminator that computes the dot product between the last discriminator layer and a learned embedding of the conditions~\cite{miyato2018cgans}. 
Additionally, the generator typically applies conditional normalization in each layer with condition-specific, learned scale and shift parameters~\cite{devries2017modulating}. 
The conditional StyleGAN2 architecture also incorporates a projection-based discriminator and conditional normalization in the generator.

Less attention has been given to multi-conditional GANs, where the conditioning is made up of multiple distinct categories of conditions that apply to each sample. Yildirim~\etal hand-crafted loss functions for different parts of the conditioning, such as shape, color, or texture on a fashion dataset~\cite{yildirim2018disentangling}.
Park~\etal proposed a GAN conditioned on a base image and a textual \emph{editing instruction} to generate the corresponding edited image~\cite{park2018mcgan}.

\paragraph{Art with GANs.}
There is a long history of attempts to emulate human creativity by means of AI methods such as neural networks.
Some studies focus on more practical aspects, whereas others consider  philosophical questions such as whether machines are able to create artifacts that evoke human emotions in the same way as human-created art does.
Recent developments include the work of Mohammed and Kiritchenko, who collected annotations, including perceived emotions and preference ratings, for over 4,000 artworks~\cite{mohammed2018artemo}.
Achlioptas~\etal introduced a dataset with less annotation variety, but were able to gather perceived emotions for over 80,000 paintings~\cite{achlioptas2021artemis}.
In addition, they solicited explanation utterances from the annotators about why they felt a certain emotion in response to an artwork, leading to around 455,000 annotations.
Elgammal~\etal presented a Creative Adversarial Network (CAN) architecture that is encouraged to produce more novel forms of artistic images by deviating from style norms rather than simply reproducing the target distribution~\cite{elgammal2017can}.
Liu~\etal proposed a new method to generate art images from sketches given a specific art style~\cite{liu2020sketchtoart}.
As certain paintings produced by GANs have been sold for high prices,\fnover\footnote{\url{https://www.christies.com/features/a-collaboration-between-two-artists-one-human-one-a-machine-9332-1.aspx}} McCormack~\etal raise important questions about issues such as authorship and copyrights of generated art~\cite{mccormack2019autonomy}.

\paragraph{GAN Inversion.}
GAN inversion is a rapidly growing branch of GAN research.
The objective of GAN inversion is to find a reverse mapping from a given genuine input image into the latent space of a trained GAN.
Xia~\etal provide a survey of prominent inversion methods and their applications~\cite{xia2021gan}.
In the context of StyleGAN, Abdal~\etal proposed Image2StyleGAN, which was one of the first feasible methods to invert an image into the extended latent space $W^+$ of StyleGAN~\cite{abdal2019image2stylegan}.
Karras~\etal instead opted to embed images into the smaller $W$ space so as to improve the editing quality at the cost of reconstruction~\cite{karras2020analyzing}.
With the latent code for an image, it is possible to navigate in the latent space and modify the produced image.
Applications of such latent space navigation include image manipulation~\cite{abdal2019image2stylegan, abdal2020image2stylegan, abdal2020styleflow, zhu2020indomain, shen2020interpreting, voynov2020unsupervised, xu2021generative}, image restoration~\cite{shen2020interpreting, pan2020exploiting, Ulyanov_2020, yang2021gan}, and image interpolation~\cite{abdal2020image2stylegan, Xia_2020, pan2020exploiting, nitzan2020face}.
We would like to highlight the work of Zhu~\etal, which introduced two new embedding spaces for GAN inversion, the $P$ space and the improved $P_N$ space~\cite{zhu2021improved}.
The $P$ space eliminates the skew of marginal distributions in the more widely used $W$ space.

\paragraph{GAN Evaluation.}
There are many evaluation techniques for GANs that attempt to assess the visual quality of generated images~\cite{devries19}.
Due to the different focus of each metric, there is not just one accepted definition of visual quality.
Also, many of the metrics solely focus on unconditional generation and evaluate the separability between generated images and real images, as for example the approach from Zhou \etal~\cite{zhou2019hype}. 
However, the Fréchet Inception Distance (FID) score by Heusel~\etal~\cite{heusel2018gans} has become commonly accepted and computes the distance between two distributions.
Variations of the FID such as the Fréchet Joint Distance {FJD}~\cite{devries19} and the Intra-Fréchet Inception Distance (I-FID)~\cite{takeru18} additionally enable an assessment of whether the conditioning of a GAN was successful.
This allows us to also assess desirable properties such as \textit{conditional consistency} and \textit{intra-condition diversity} of our GAN models~\cite{devries19}. 
Another frequently used metric to benchmark GANs is the Inception Score (IS)~\cite{salimans16}, which primarily considers the diversity of samples.

\section{Compiling an Annotated Dataset}
\label{sec:dataset}

\begin{table}
  \centering
  \begin{tabular}{@{}lcrc@{}}
    \toprule
    Feature & Type & Size & Example \\
    \midrule
    Style & Category & 29 & Post-Impressionism \\
    Painter & Category & 351 & Vincent van Gogh \\
    Genre & Category & 30 & cloudscape\\
    Content tags & Text & 768 & tree, sky\\
    Emotions & Distribution & 9 & 40\% awe, ...\\
    Utterance & Text & 768 & ``The sky seems ..."\\
    \bottomrule
  \end{tabular}
  \caption{Features in the EnrichedArtEmis dataset, with example values  for ``The Starry Night" by Vincent van Gogh.}
  \label{tab:labels-art-dataset}
\end{table}

WikiArt\footnote{\url{https://www.wikiart.org/}} is an online encyclopedia of visual art that catalogs both historic and more recent artworks. Similar to Wikipedia, the service accepts community contributions and is run as a non-profit endeavor.

The ArtEmis dataset~\cite{achlioptas2021artemis} contains roughly 80,000 artworks obtained from WikiArt, enriched with additional human-provided emotion annotations.
On average, each artwork has been annotated by six different non-expert annotators with one out of nine possible emotions (\emph{amusement}, \emph{awe}, \emph{contentment}, \emph{excitement}, \emph{disgust}, \emph{fear}, \emph{sadness}, \emph{other}) along with a sentence (utterance) that explains their choice.

We enhance this dataset by adding further metadata crawled from the WikiArt website -- \emph{genre}, \emph{style}, \emph{painter}, and \emph{content} tags -- that serve as conditions for our model. 
When a particular attribute is not provided by the corresponding WikiArt page, we assign it a special \textsc{Unknown} token. 
Additionally, in order to reduce issues introduced by conditions with low support in the training data, we also replace all categorical conditions that appear less than 100 times with this \textsc{Unknown} token. 
We refer to this enhanced version as the \textbf{EnrichedArtEmis} dataset.

A summary of the conditions present in the EnrichedArtEmis dataset is given in \autoref{tab:labels-art-dataset}. 
Note that our conditions have different modalities. 
The conditions \emph{painter}, \emph{style}, and \emph{genre} are categorical and encoded using one-hot encoding. Emotion annotations are provided as a discrete probability distribution over the respective emotion labels, as there are multiple annotators per image, i.e., each element denotes the percentage of annotators that labeled the corresponding choice for an image.
Finally, we have textual conditions, such as content tags and the annotator explanations from the ArtEmis dataset. 
For these, we use a pretrained TinyBERT model to obtain 768-dimensional embeddings \cite{jiao2020tinybert}.

\section{Exploring Conditional StyleGAN}
\label{sec:exploration}

The StyleGAN architecture~\cite{karras2019stylebased} introduced by Karras~\etal and the improved version StyleGAN2~\cite{karras2020analyzing} produce images of good quality and high resolution.
With an adaptive augmentation mechanism, Karras~\etal were able to reduce the data and thereby the cost needed to train a GAN successfully~\cite{karras2020training}.
For brevity, in the following, we will refer to StyleGAN2-ADA, which includes the revised architecture and the improved training, as StyleGAN.

The StyleGAN architecture consists of a mapping network and a synthesis network.
Given a latent vector $\vec{z}$ in the input latent space $Z$, the non-linear mapping network $f : Z \rightarrow W$ produces $\wvec \in W$.
The mapping network is used to disentangle the latent space $Z$.
The latent vector $\wvec$ then undergoes some modifications when fed into every layer of the synthesis network to produce the final image.
This architecture improves the understanding of the generated image, as the synthesis network can distinguish between coarse and fine features.
In addition, it enables new applications, such as style-mixing, where two latent vectors from $W$ are used in different layers in the synthesis network to produce a mix of these vectors.

In order to influence the images created by networks of the GAN architecture, a conditional GAN (cGAN) was introduced by Mirza and Osindero~\cite{mirza2014conditional} shortly after the original introduction of GANs by Goodfellow~\etal~\cite{goodfellow2014generative}.
In their work, Mirza and Osindera simply fed the conditions alongside the random input vector and were able to produce images that fit the conditions.
Over time, more refined conditioning techniques were developed, such as an auxiliary classification head in the discriminator~\cite{odena2017conditional} and a projection-based discriminator~\cite{miyato2018cgans}.
The StyleGAN generator follows the approach of accepting the conditions as additional inputs but uses conditional normalization in each layer with condition-specific, learned scale and shift parameters~\cite{devries2017modulating, karras-stylegan2}.

With a latent code $\vec{z}$ from the input latent space $Z$ and a condition $\vec{c}$ from the condition space $C$, the non-linear conditional mapping network $\fc : Z, C \rightarrow W$ produces $\wc \in W$.
The latent code $\wc$ is then used together with conditional normalization layers in the synthesis network of the generator to produce the image.
The discriminator uses a projection-based conditioning mechanism~\cite{miyato2018cgans, karras-stylegan2}.
In the following, we study the effects of conditioning a StyleGAN.

\subsection{Choice of Space to Evaluate Conditioning}

\begin{figure}[t]
  \centering
   \includegraphics[width=1.0\linewidth]{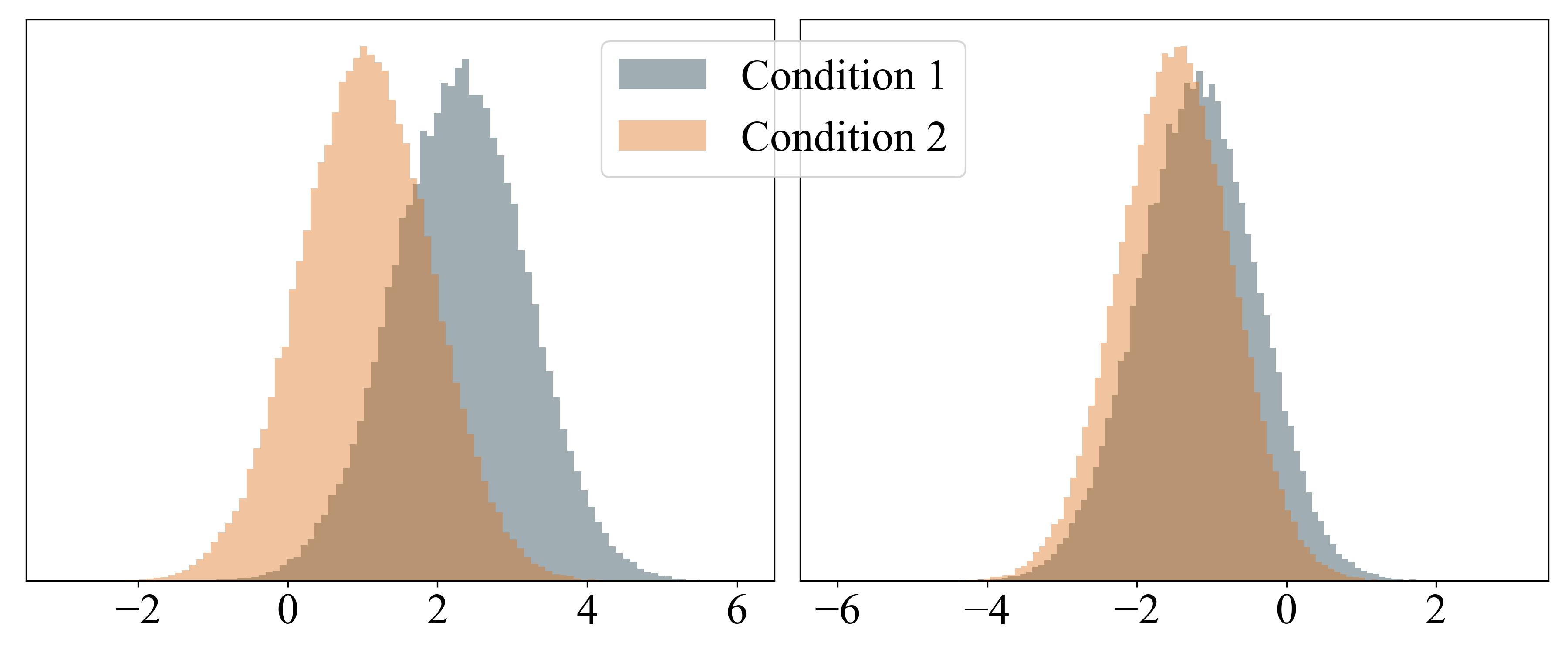}
   \caption{Samples in $P$ space for two conditions in two randomly chosen dimensions. As the distributions for the conditions look different, we believe they will behave differently as well.}
   \label{fig:p-distribution}
\end{figure}

The most obvious way to investigate the conditioning is to look at the images produced by the StyleGAN generator.
However, this is highly inefficient, as generating thousands of images is costly and we would need another network to analyze the images.
The StyleGAN architecture and in particular the mapping network is very powerful.
The mapping network, an 8-layer MLP, is not only used to disentangle the latent space, but also embeds useful information about the condition space.
Hence, we consider a condition space before the synthesis network as a suitable means to investigate the conditioning of the StyleGAN.
Analyzing an embedding space before the synthesis network is much more cost-efficient, as it can be analyzed without the need to generate images. 

As explained in the survey on GAN inversion by Xia~\etal, a large number of different embedding spaces in the StyleGAN generator may be considered for successful GAN inversion~\cite{xia2021gan}.
Usually these spaces are used to embed a given image back into StyleGAN. Thus, all kinds of modifications, such as image manipulation~\cite{abdal2019image2stylegan, abdal2020image2stylegan, abdal2020styleflow, zhu2020indomain, shen2020interpreting, voynov2020unsupervised, xu2021generative}, image restoration~\cite{shen2020interpreting, pan2020exploiting, Ulyanov_2020, yang2021gan}, and image interpolation~\cite{abdal2020image2stylegan, Xia_2020, pan2020exploiting, nitzan2020face} can be applied.
An obvious choice would be the aforementioned $W$ space, as it is the output of the mapping network.
However, Zhu~\etal discovered that ``the marginal distributions [in $W$] are heavily skewed and do not follow an obvious pattern"~\cite{zhu2021improved}.
They therefore proposed the $P$ space and building on that the $P_N$ space.
The $P$ space can be obtained by inverting the last LeakyReLU activation function in the mapping network that would normally produce the $W$ space.
The last LeakyReLU has a slope of $\alpha = 0.2$ and we can thus transform the $W$ space to the $P$ space with:

\begin{equation}
  \vec{x} = \text{LeakyReLU}_{5.0}(\wvec)
  \label{eq:p-space}
\end{equation}

\noindent where $\wvec$ and $\vec{x}$ are vectors in the latent spaces $W$ and $P$, respectively.
Zhu~\etal make the assumption that the joint distribution of points in the latent space $P$ approximately follow a multivariate Gaussian distribution~\cite{zhu2021improved}.
To examine whether the $P$ space is a valid embedding space to evaluate the conditioning, we train a conditional StyleGAN on our EnrichedArtEmis data and visualize the marginal distributions of two different conditions in the $P$ space, which can be seen in \autoref{fig:p-distribution}.
We observe that the distributions for different conditions differ and that the $P$ space appears to follow a Gaussian distribution.
To support our assumption that the distributions for different conditions are indeed different, we compare their respective multivariate Gaussian distributions.
Hence, we do not need to invoke the $P_N$ space, whose main purpose is to eliminate the dependency among latent variables, as the multivariate Gaussian distribution will address these dependencies.

\subsection{Analyzing the Conditioning in StyleGAN}
\label{sec:exploration-distribution}

\begin{figure}[t]
  \centering
  \begin{subfigure}{0.32\linewidth}
    \includegraphics[width=1\linewidth]{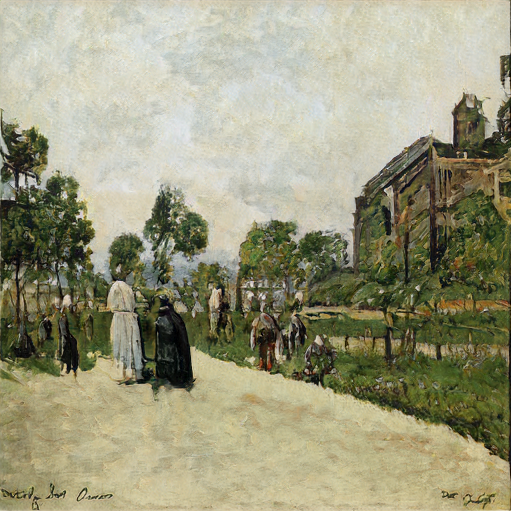}
    \caption*{Impressionism}
  \end{subfigure}
  \hfill
\begin{subfigure}{0.32\linewidth}
    \includegraphics[width=1\linewidth]{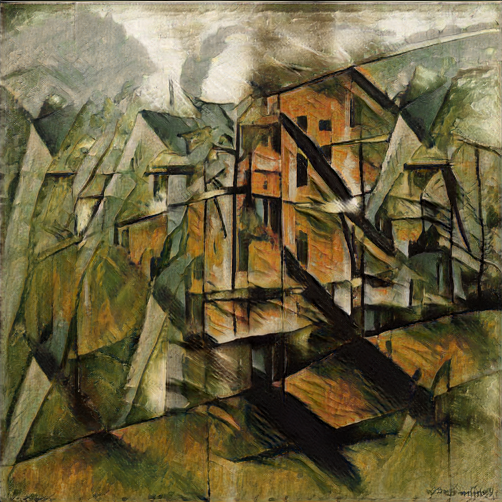}
    \caption*{Cubism}
  \end{subfigure}
  \hfill
  \begin{subfigure}{0.32\linewidth}
    \includegraphics[width=1\linewidth]{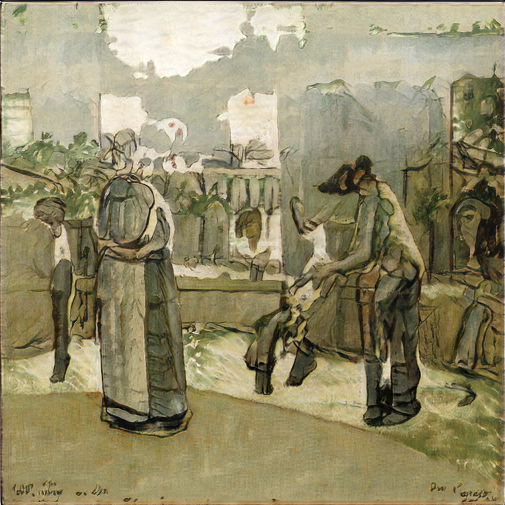}
    \caption*{Expressionism}
  \end{subfigure}
  \caption{Paintings produced by a StyleGAN model conditioned on style. All images are generated with identical random noise $\zvec$, yielding images with a similar color palette.}
  \label{fig:style}
\end{figure}

For each condition $c$, we sample 10,000 points in the latent $P$ space: $X_c \in \mathbb{R}^{10^4 \times n}$.
The $P$ space has the same size as the $W$ space with $n = 512$.
We determine mean $\muvec_c \in \mathbb{R}^n$ and covariance matrix $\Sigma_c$ for each condition $c$ based on the samples $X_c$. For each condition $c$, we obtain a multivariate normal distribution $\mathcal{N}(\muvec_c, \Sigma_c)$.

We create 100,000 additional samples $Y_c \in \mathbb{R}^{10^5 \times n}$ in $P$ for each condition.
We wish to predict the label of these samples based on the given multivariate normal distributions.
The probability that a vector $\vec{x} \in \mathbb{R}^n$ belongs to a multivariate normal distribution with mean $\muvec \in \mathbb{R}^n$ and covariance matrix $\Sigma \in \mathbb{R}^{n \times n}$ is defined by the probability density function of the multivariate Gaussian distribution:
\begin{equation}
  p(\vec{x};\muvec,\Sigma) = \frac{1}{(2\pi)^{n/2}|\Sigma|^{1/2}} \; e^{-\frac{1}{2} (\vec{x}-\muvec)^\intercal \Sigma^{-1} (\vec{x}-\muvec)}
  \label{eq:gaussian-pdf}
\end{equation}
The condition $\hat{c}$ we assign to a vector $\vec{x} \in \mathbb{R}^n$ is defined as the condition that achieves the highest probability score based on the probability density function (\autoref{eq:gaussian-pdf}), i.e.,
\begin{equation}
  \hat{c} = \argmax_{c \, \in \, C} \; p(\vec{x};\muvec_c,\Sigma_c)
  \label{eq:gaussian-label}
\end{equation}
Having trained a StyleGAN model on the EnrichedArtEmis dataset, 
we find that we are able to assign \emph{every} vector $\vec{x} \in Y_c$ the correct label $c$.
This strengthens the assumption that the distributions for different conditions are indeed different.

\subsection{Similarity Between Classes}

\begin{table} \centering
    \begin{tabular}{@{} l*{5}r @{}}
        & \rot{Baroque} & \rot{Rococo} & \rot{\shortstack[l]{High\\Renaissance}} & \rot{Minimalism} & \rot{\shortstack[l]{Color Field\\Painting}}\\
        \cmidrule{1-6}
        Baroque & 0.0 & \textbf{10.4} & 11.1 & 16.7 & 19.2\\
        Rococo & \textbf{10.4} & 0.0 & 13.7 & 18.3 & 20.3\\
        High Renaissance & \textbf{11.1} & 13.7 & 0.0 & 17.0 & 19.6\\
        Minimalism & 16.7 & 18.3 & 17.0 & 0.0 & \textbf{10.9}\\
        Color Field Painting & 19.2 & 20.3 & 19.6 & \textbf{10.9} & 0.0\\
        \cmidrule{1-6}
    \end{tabular}
    \caption{Fréchet distances for selected art styles. For each art style the lowest FD to an art style other than itself is marked in bold.}
    \label{tab:style-fd}
\end{table}

\begin{figure}[t]
  \centering
  \includegraphics[width=1\linewidth]{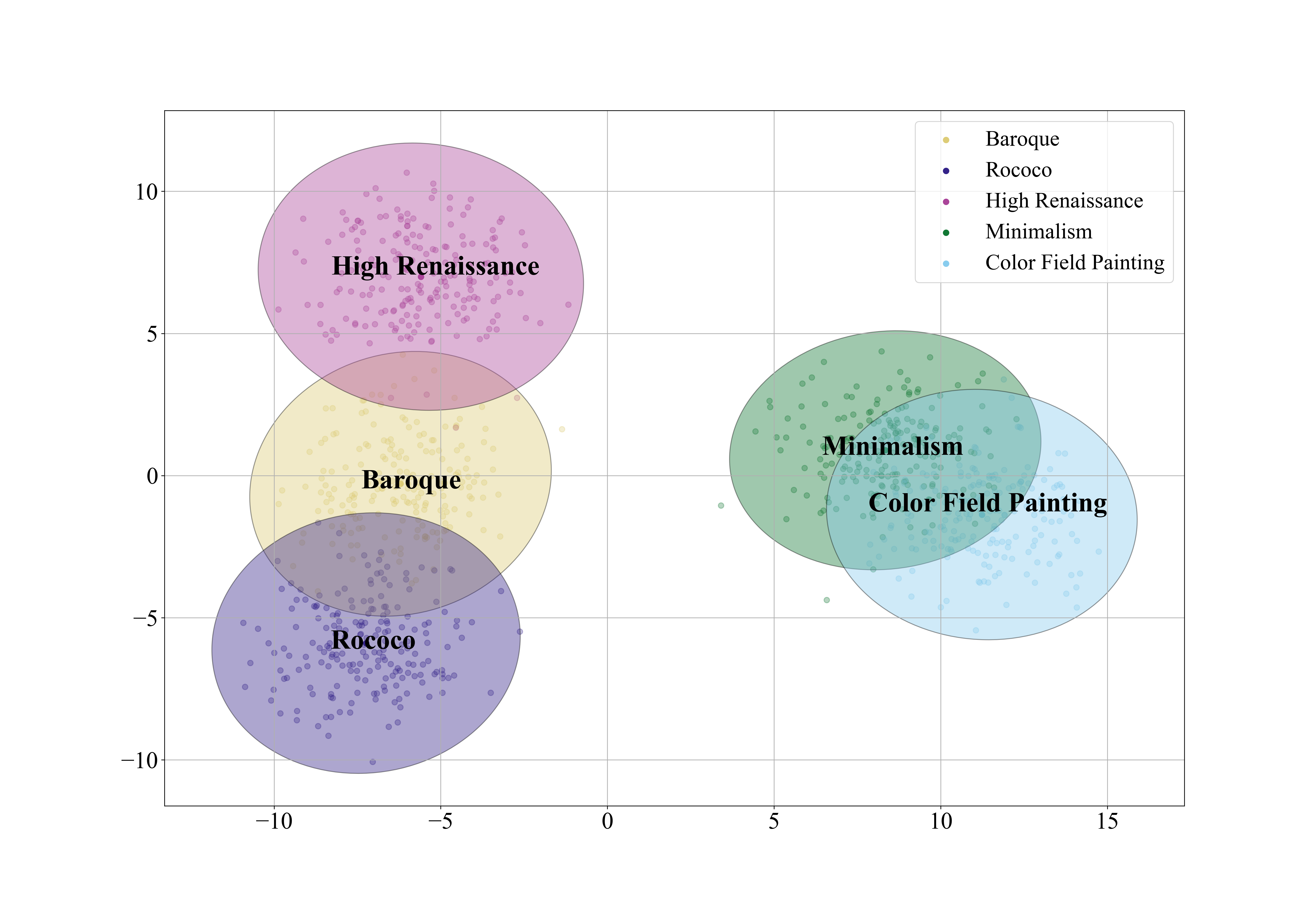}
   \caption{Dimensionality reduction with PCA of the $P$ space. Shown are the art styles \emph{High Renaissance}, \emph{Baroque}, \emph{Rococo}, \emph{Color Field Painting}, \emph{Minimalism}. Highlighted regions denote the covariance confidence with $3\sigma$ for each art style.}
   \label{fig:simplified-pca}
\end{figure}

Now that we know that the $P$ space distributions for different conditions behave differently, we wish to analyze these distributions.
We can compare the multivariate normal distributions and investigate similarities between conditions.
To this end, we use the Fréchet distance (FD) between multivariate Gaussian distributions~\cite{dowson1982frechet}:
\begin{equation}
  \text{FD}^2 = ||\muvec_{c_1} - \muvec_{c_2}||^2 + \Tr(\Sigma_{c_1} + \Sigma_{c_2} - 2\sqrt{\Sigma_{c_1} \Sigma_{c_2}})
  \label{eq:fd}
\end{equation}
where $X_{c_1} \sim \mathcal{N}(\muvec_{c_1}, \Sigma_{c_1})$ and $X_{c_2} \sim \mathcal{N}(\muvec_{c_2}, \Sigma_{c_2})$ are distributions from the $P$ space for conditions $c_1, c_2 \in C$.
The lower the FD between two distributions, the more similar the two distributions are and the more similar the two conditions that these distributions are sampled from are, respectively.
The most well-known use of FD scores is as a key component of Fréchet Inception Distance (FID)~\cite{heusel2018gans}, which is used to assess the quality of images generated by a GAN. In that setting, the FD is applied to the 2048-dimensional output of the Inception-v3~\cite{szegedy2015rethinking} pool3 layer for real and generated images.

\subsection{Use Cases of Conditioning Space}

We train a StyleGAN on the paintings in the EnrichedArtEmis dataset, which contains around 80,000 paintings from 29 art styles, such as impressionism, cubism, expressionism, etc.~\cite{achlioptas2021artemis}.
We condition the StyleGAN on these art styles to obtain a conditional StyleGAN.
Examples of generated images can be seen in \autoref{fig:style}.
We compute the FD for all combinations of distributions in $P$ based on the StyleGAN conditioned on the art style.
The FDs for a selected number of art styles are given in \autoref{tab:style-fd}. The obtained FD scores
suggest a high degree of similarity between the art styles Baroque, Rococo, and High Renaissance.
Furthermore, the art styles Minimalism and Color Field Painting seem similar. From an art historic perspective, these clusters indeed appear reasonable.

Another application is the visualization of differences in art styles.
For this, we use Principal Component Analysis (PCA) on $X_c$ to reduce the $P$ space from $512$ to two dimensions.
We do this for the five aforementioned art styles and keep an explained variance ratio of nearly 20\%.
The results are visualized in \autoref{fig:simplified-pca}.
We can observe that the results accord with the clustering obtained with FD. In addition, it appears that Baroque ends up residing between High Renaissance and Rococo.
Historically, the Baroque period indeed lies between the High Renaissance and Rococo and their themes match accordingly.

\subsection{Predicting Inverted Images}

We have shown that it is possible to predict a latent vector sampled from the latent space $Z$.
However, we can also apply GAN inversion to further analyze the latent spaces.
GAN inversion seeks to map a real image into the latent space of a pretrained GAN.
We believe it is possible to invert an image and predict the latent vector according to the method from Section \ref{sec:exploration-distribution}.
While this operation is too cost-intensive to be applied to large numbers of images, it can simplify the navigation in the latent spaces if the initial position of an image in the respective space can be assigned to a known condition.

\section{Methods for Conditional StyleGANs}
\label{sec:conditional-generation}
In this section, we investigate two methods that use conditions in the $W$ space to improve the image generation process. We recall our definition for the unconditional mapping network: a non-linear function $f : Z \rightarrow W$ that maps a latent code $\zvec \in Z$ to a latent vector $\wvec \in W$. For conditional generation, the mapping network is extended with the specified conditioning $c \in C$ as an additional input to $\fc : Z, C \rightarrow W$.

\subsection{Conditional Truncation}
\label{sec:conditional-trunc}
\begin{figure}
  \centering
  \begin{subfigure}{0.48\linewidth}
    \includegraphics[width=1\linewidth]{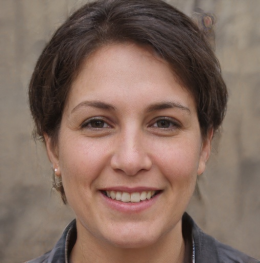}
    \caption{Image produced by the center of mass on FFHQ. Taken from Karras~\etal~\cite{karras2019stylebased}.}
    \label{fig:com-ffhq}
  \end{subfigure}
  \hfill
  \begin{subfigure}{0.48\linewidth}
    \includegraphics[width=1\linewidth]{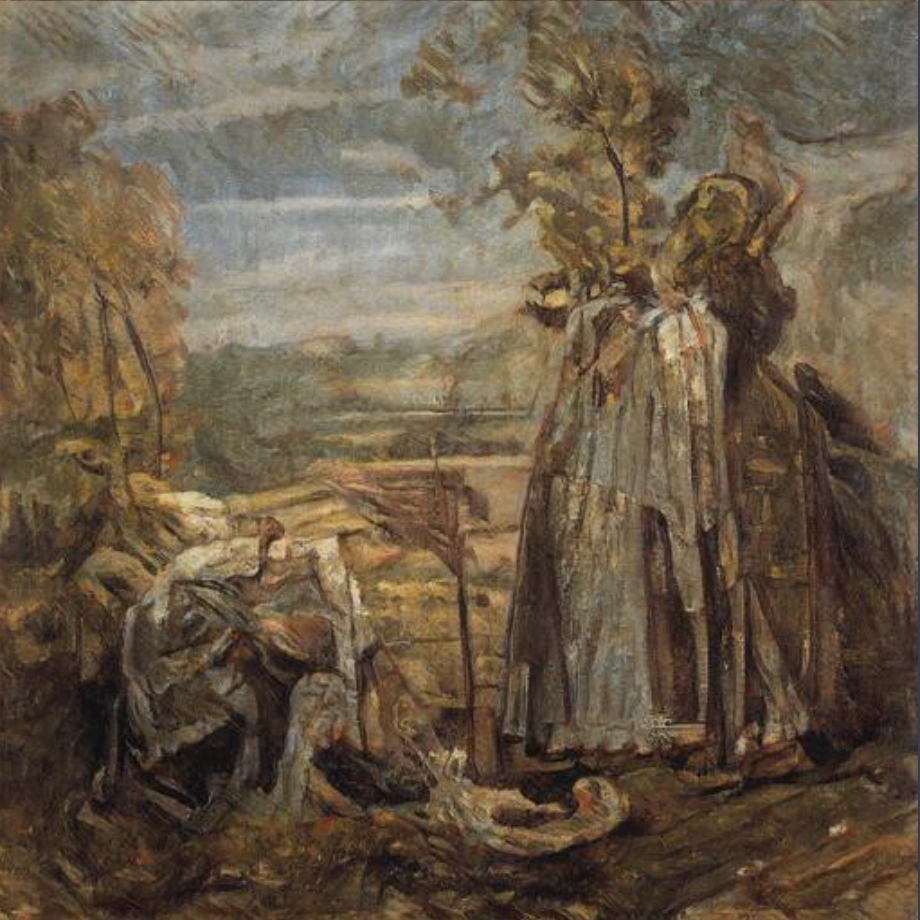}
    \caption{Image produced by the center of mass on EnrichedArtEmis.\\~}
    \label{fig:com-our}
  \end{subfigure}
  \caption{Images produced by center of masses for StyleGAN models that have been trained on different datasets.}
  \label{fig:com}
\end{figure}

\begin{figure*}[htbp!]
\centering 
     \begin{subfigure}{0.18\textwidth}
        \includegraphics[width=\linewidth]{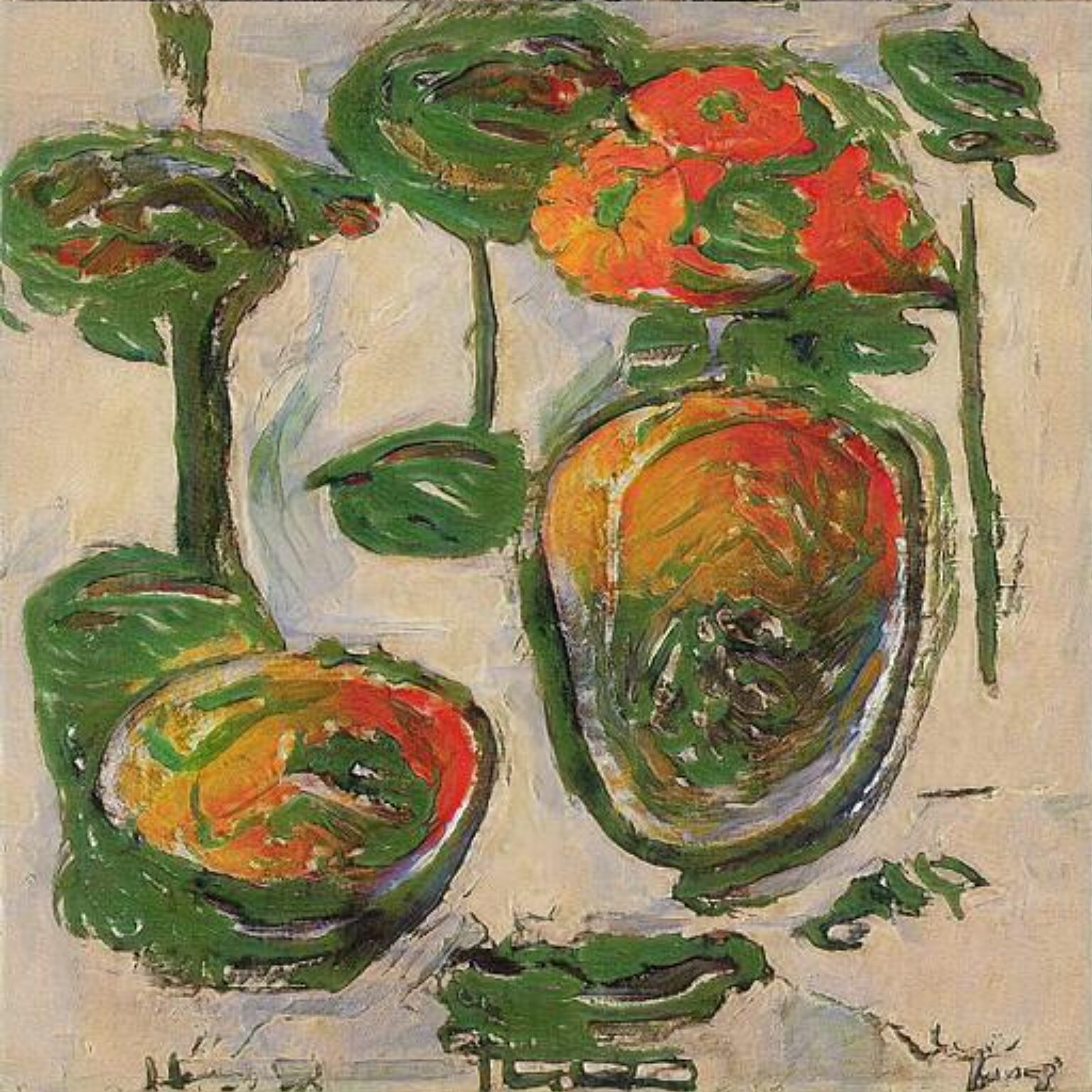}
        \caption*{$\psi = 1$}
      \end{subfigure}
    \begin{subfigure}{0.18\textwidth}
        \includegraphics[width=\linewidth]{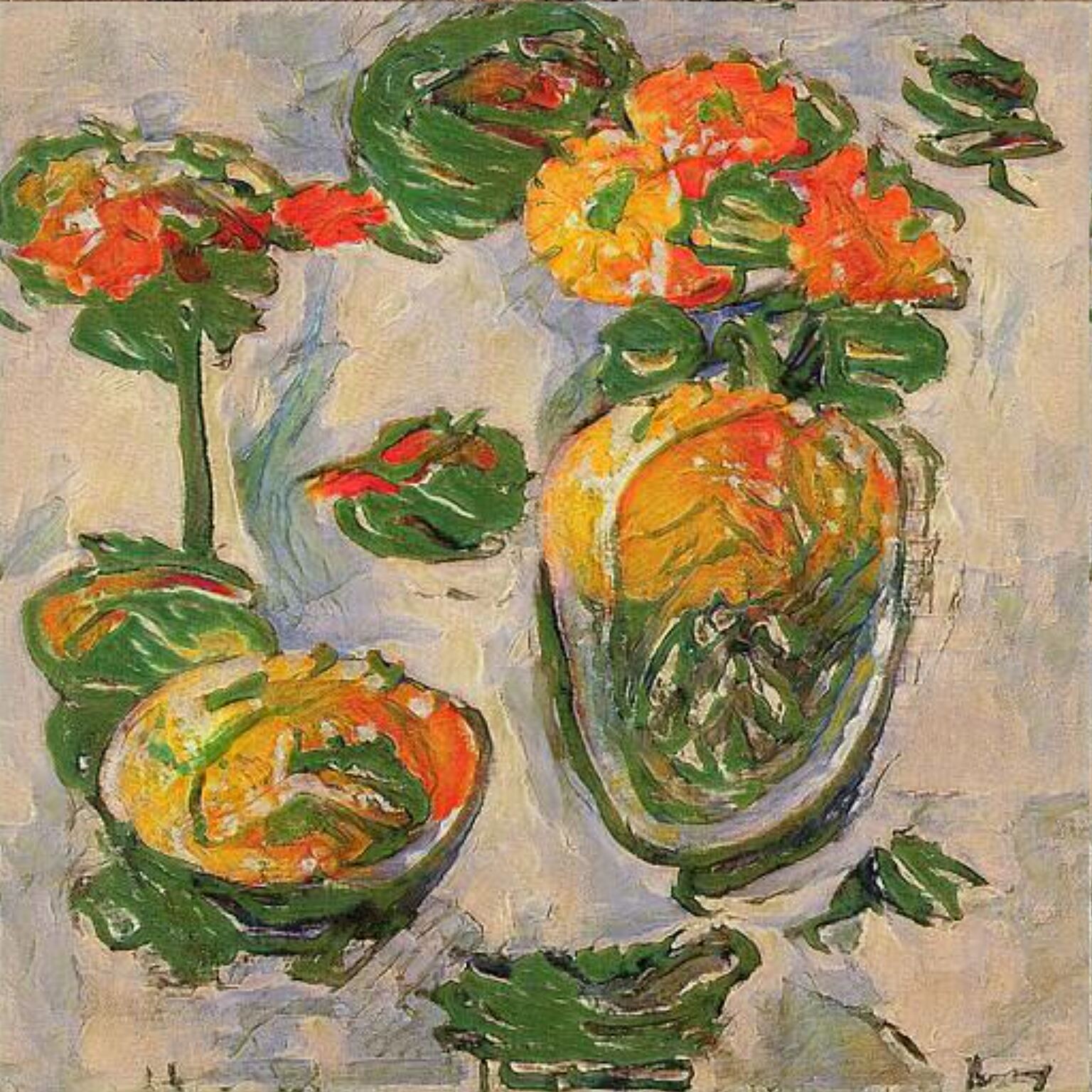}
        \caption*{$\psi = 0.75$}

    \end{subfigure}
    \begin{subfigure}{0.18\textwidth}
        \includegraphics[width=\linewidth]{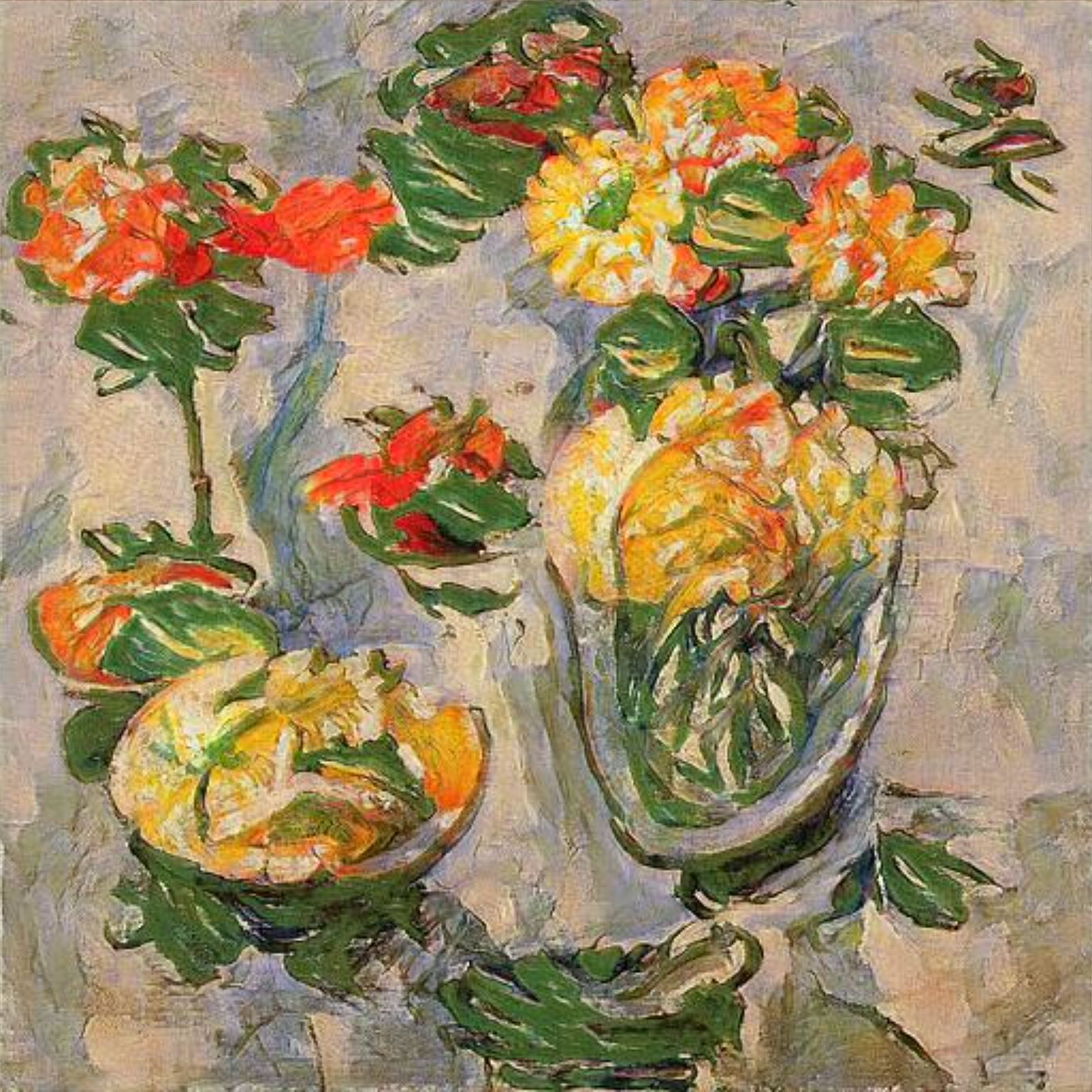}
        \caption*{$\psi = 0.5$}
    \end{subfigure}
    \begin{subfigure}{0.18\textwidth}
        \includegraphics[width=\linewidth]{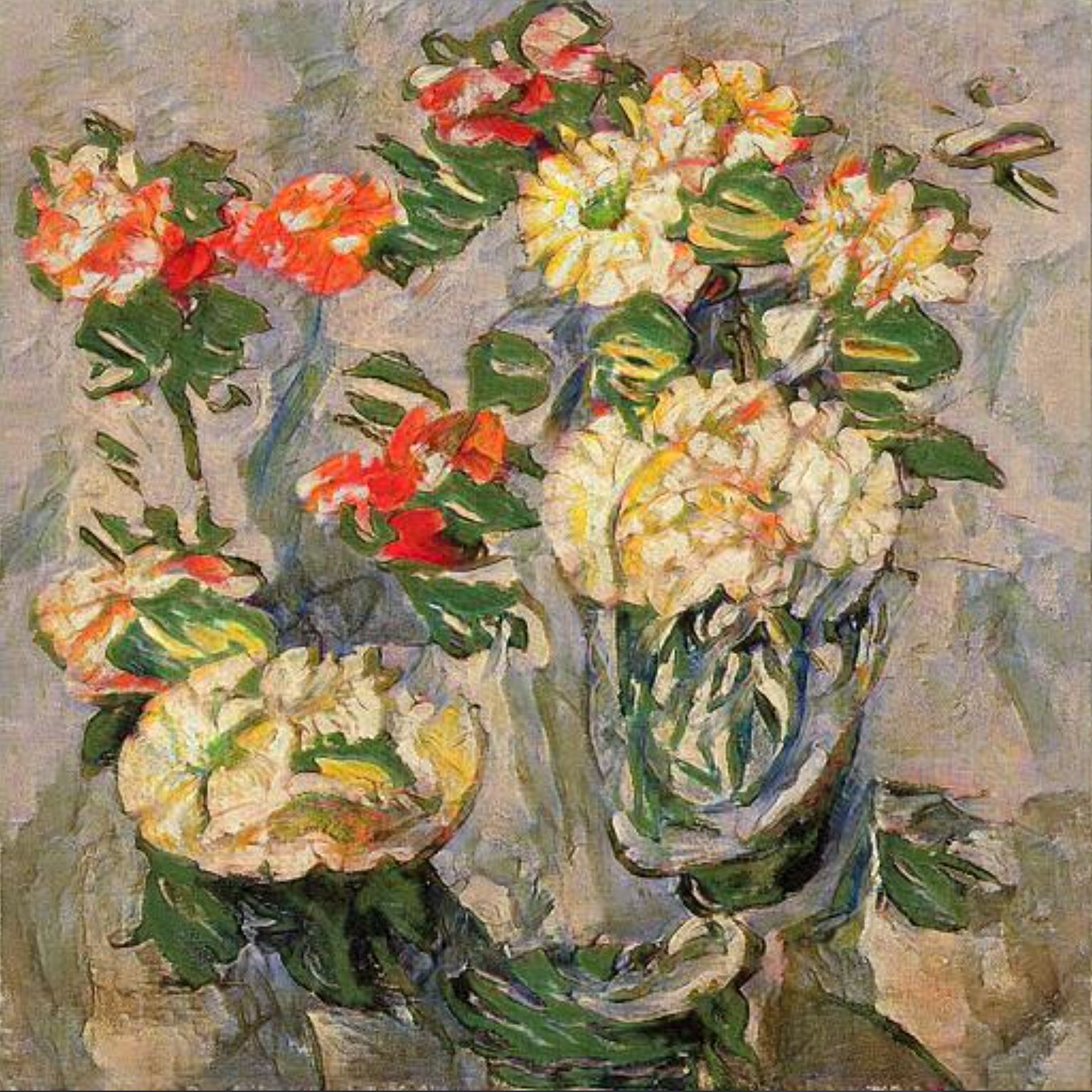}
        \caption*{$\psi = 0.25$}
    \end{subfigure}
    \begin{subfigure}{0.18\textwidth}
        \includegraphics[width=\linewidth]{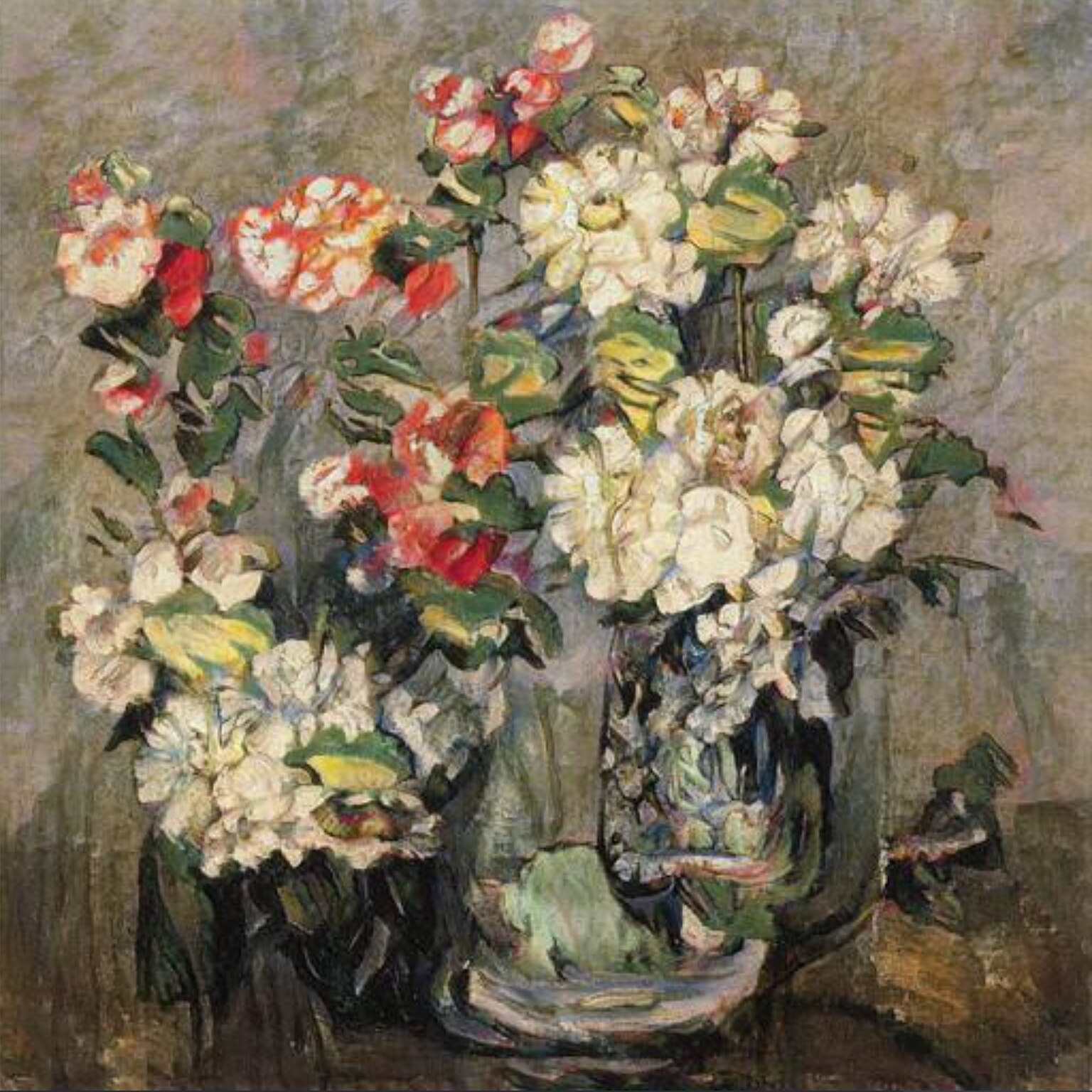}
        \caption*{$\psi = 0$}
    \end{subfigure}
    \captionsetup{width=.9\textwidth}
    \caption{Visualization of the conditional truncation trick with the condition \textit{flower paintings}. As $\psi \to 0$ and the sampled $\wvec_c$ is moved towards the \textit{conditional} center of mass, the \textit{flower painting} condition is retained. Also, the \textit{conditional} center of mass at $\psi=0$ produces a high-fidelity image.}
    \label{fig:conditional-truncation-trick}
\end{figure*}

\begin{figure*}[htbp!]
\centering 
     \begin{subfigure}{0.18\textwidth}
        \includegraphics[width=\linewidth]{figures/truncation/original}
        \caption*{$\psi = 1$}
      \end{subfigure}
    \begin{subfigure}{0.18\textwidth}
        \includegraphics[width=\linewidth]{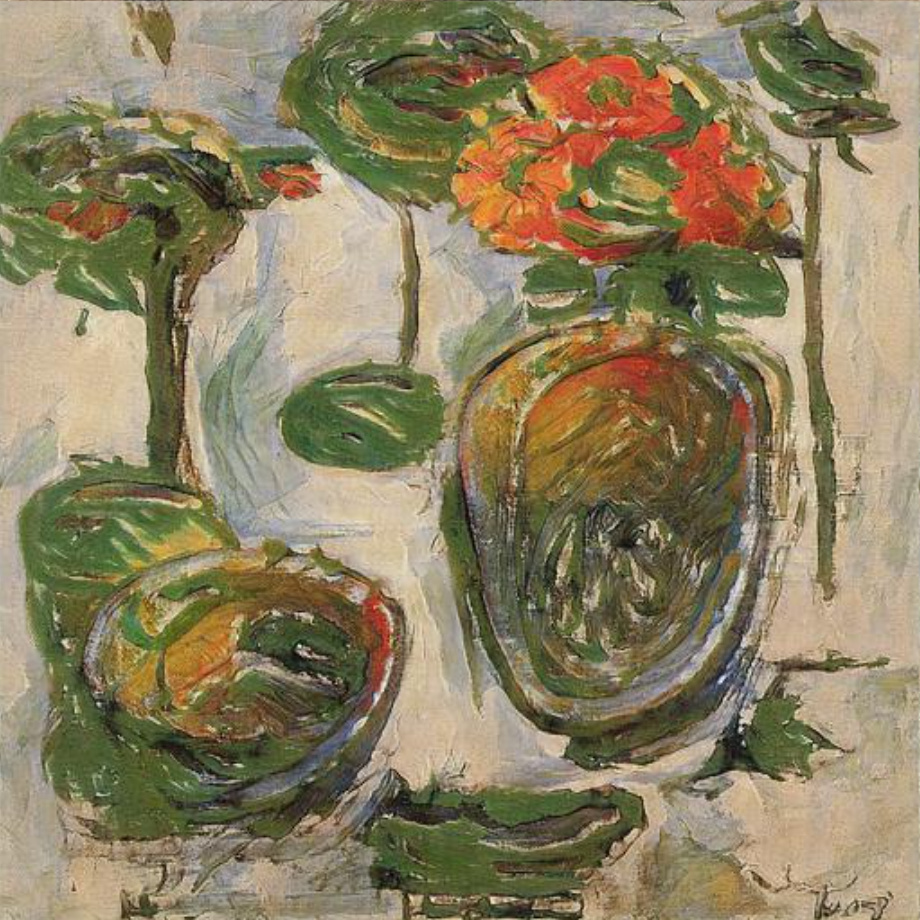}
        \caption*{$\psi = 0.75$}

    \end{subfigure}
    \begin{subfigure}{0.18\textwidth}
        \includegraphics[width=\linewidth]{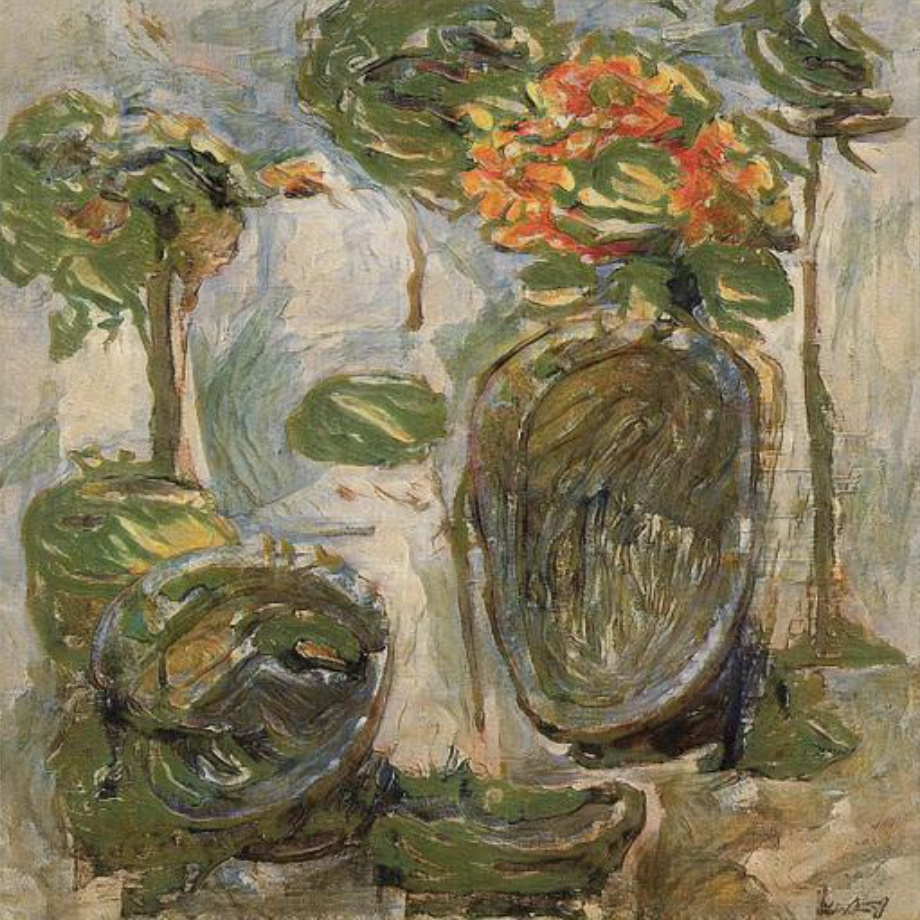}
        \caption*{$\psi = 0.5$}
    \end{subfigure}
    \begin{subfigure}{0.18\textwidth}
        \includegraphics[width=\linewidth]{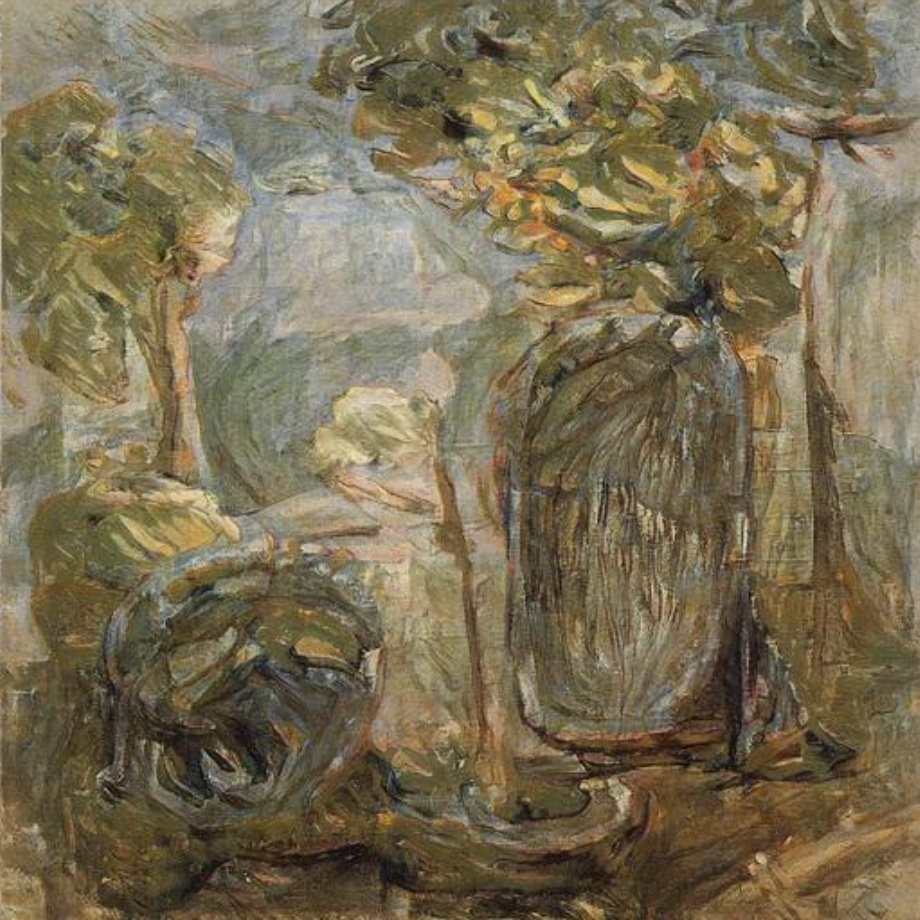}
        \caption*{$\psi = 0.25$}
    \end{subfigure}
    \begin{subfigure}{0.18\textwidth}
        \includegraphics[width=\linewidth]{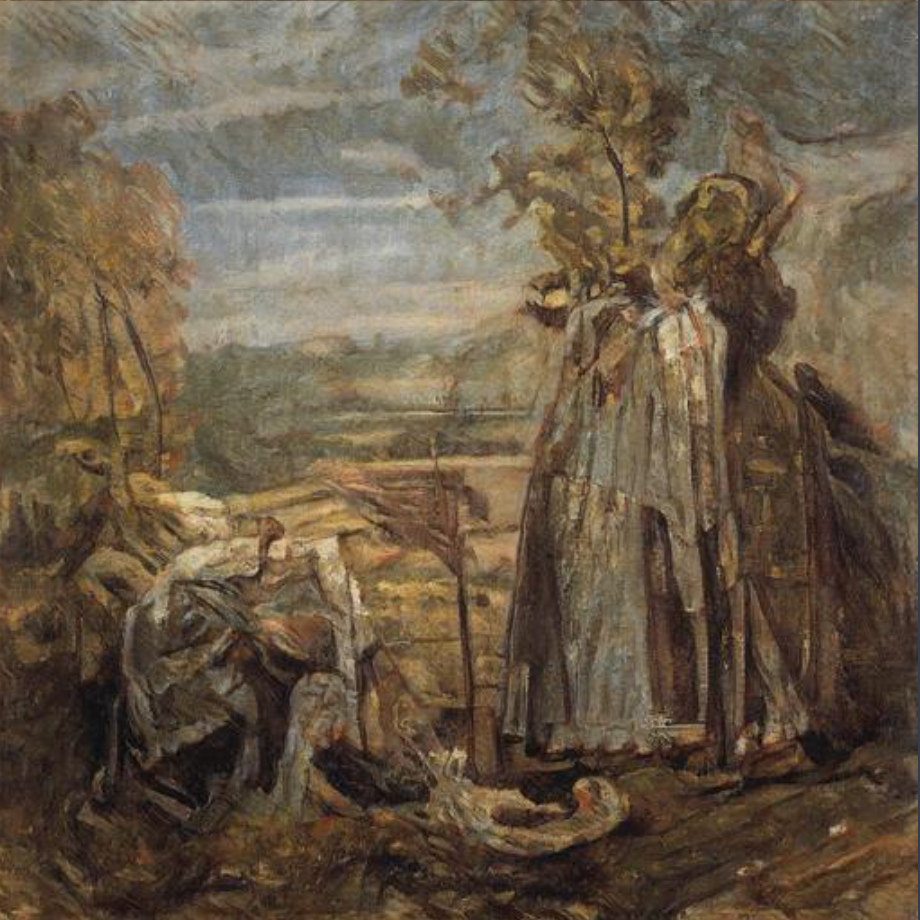}
        \caption*{$\psi = 0$}
    \end{subfigure}
    \captionsetup{width=.9\textwidth}
    \caption{Visualization of the conventional truncation trick with the condition \textit{flower paintings}. As $\psi \to 0$ and the sampled $\wvec_c$ is moved towards the \textit{global} center of mass, the \textit{flower painting} condition is increasingly lost. Moreover, the \textit{global} center of mass at $\psi=0$ yields a low-fidelity image.}
    \label{fig:normal-truncation-trick}
\end{figure*}

The \textit{truncation trick}~\cite{brock2018largescalegan} is a method to adjust the tradeoff between the fidelity (to the training distribution) and diversity of generated images by truncating the space from which latent vectors are sampled. 
Based on its adaptation to the StyleGAN architecture by Karras~\etal~\cite{karras2019stylebased}, we propose a variant of the truncation trick specifically for the conditional setting. 

For the StyleGAN architecture, the truncation trick works by first computing the \textit{global} center of mass in $W$ as
\begin{align}
    \bar{\wvec} = \mathbb{E}_{\zvec \sim P(\zvec)}[f(\zvec)].
\end{align}

Then, a given sampled vector $\wvec$ in $W$ is moved towards $\bar{\wvec}$ with
\begin{align}
    \wvec' = \bar{\wvec} + \psi (\wvec - \bar{\wvec}),
    \label{eq:truncation-move-to-center}
\end{align}
\noindent where $\psi < 1$. 

Moving towards a global center of mass has two disadvantages: Firstly, the \textit{condition retention} problem, where the conditioning of an image is lost progressively the more we apply the truncation trick. 
The reason is that the image produced by the global center of mass in $W$ does not adhere to any given condition. 
The more we apply the truncation trick and move towards this global center of mass, the more the generated samples will deviate from their originally specified condition.
In the conditional setting, adherence to the specified condition is crucial and deviations can be seen as detrimental to the quality of an image.
Therefore, the conventional truncation trick for the StyleGAN architecture is not well-suited for our setting.

Secondly, when dealing with datasets with structurally diverse samples, such as EnrichedArtEmis, the global center of mass itself is unlikely to correspond to a high-fidelity image. 
For the Flickr-Faces-HQ (FFHQ) dataset by Karras~\etal~\cite{karras2019stylebased}, the global center of mass produces a ``typical", high-fidelity face (\autoref{fig:com-ffhq}). 
The FFHQ dataset contains centered, aligned and cropped images of faces and therefore has low structural diversity. 
On EnrichedArtEmis however, the global center of mass does not produce a high-fidelity painting (see \autoref{fig:com-our}).
We believe this is because there are no structural patterns that govern what an art painting looks like, leading to high structural diversity. 
Therefore, as we move towards this low-fidelity global center of mass, the sample will also decrease in fidelity. Hence, applying the truncation trick is counterproductive with regard to the originally sought tradeoff between fidelity and the diversity.

Instead, we propose the \textit{conditional} truncation trick,  based on the intuition that different conditions are bound to have different centers of mass in $W$.
The mean of a set of randomly sampled $\wvec$ vectors of flower paintings is going to be different than the mean of randomly sampled $\wvec$ vectors of landscape paintings.
Thus, we compute a separate \textit{conditional} center of mass $\bar{\wvec}_c$ for each condition $c$:

\begin{equation}
  \bar{\wvec}_c = \mathbb{E}_{\zvec \sim P(\zvec)}[\fc(\zvec, c)].
  \label{eq:w-truncation}
\end{equation}

The computation of $\bar{\wvec}_c$ involves only the \textit{mapping network} and not the bigger synthesis network. 
This enables an ``on-the-fly" computation of $\bar{\wvec}_c$ at inference time for a given condition $c$. 
Moving a given vector $\wvec$ towards a conditional center of mass is done analogously to \autoref{eq:truncation-move-to-center}:
\begin{align}
    \wvec' = \bar{\wvec}_c + \psi (\wvec - \bar{\wvec}_c)
\end{align}
\noindent with $\psi < 1$.

We find that the introduction of a conditional center of mass is able to alleviate both the condition retention problem as well as the problem of low-fidelity centers of mass. 
Naturally, the conditional center of mass for a given condition will adhere to that specified condition. 
Therefore, as we move towards that conditional center of mass, we do not lose the conditional adherence of generated samples. 
In contrast, the closer we get towards the conditional center of mass, the more the conditional adherence will increase.
This effect of the conditional truncation trick can be seen in \autoref{fig:conditional-truncation-trick}, where the flower painting condition is reinforced the closer we move towards the conditional center of mass.
When using the standard truncation trick, the condition is progressively lost, as can be seen in \autoref{fig:normal-truncation-trick}.

Also, for datasets with low intra-class diversity, samples for a given condition have a lower degree of structural diversity. 
Although there are no \textit{universally applicable} structural patterns for art paintings, there certainly are \textit{conditionally applicable} patterns.
For example, flower paintings usually exhibit flower petals.
On diverse datasets that nevertheless exhibit low intra-class diversity, a conditional center of mass is therefore more likely to correspond to a high-fidelity image than the global center of mass. This effect can be observed in Figures~\ref{fig:conditional-truncation-trick} and \ref{fig:normal-truncation-trick} when considering the centers of mass with $\psi=0$.

\subsection{Condition-Based Vector Arithmetic}
\label{sec:vec-arithmetic}

Given a trained conditional model, we can steer the image generation process in a specific direction. However, it is possible to take this even further. 
As we have a latent vector $\wvec$ in $W$ corresponding to a generated image, we can apply transformations to $\wvec$ in order to alter the resulting image. 
One such transformation is vector arithmetic based on conditions: what transformation do we need to apply to $\wvec$ to change its conditioning?

Let $\wvec_{c_1}$ be a latent vector in $W$ produced by the mapping network. 
The inputs are the specified condition $c_1 \in C$ and a random noise vector $\zvec$.
Furthermore, let $\wvec_{c_2}$ be another latent vector in $W$ produced by the same noise vector but with a different condition $c_2 \neq c_1$. 
We seek a transformation vector $\vec{t}_{c_1,c_2}$ such that $\wvec_{c_1} + \vec{t}_{c_1,c_2} \approx \wvec_{c_2}$. 
Rather than just applying to a specific combination of $\zvec \in Z$ and $c_1 \in C$, this transformation vector should be generally applicable. 
Hence, we attempt to find the \textit{average} difference between the conditions $c_1$ and $c_2$ in the $W$ space.

\begin{figure*}[!htbp]
\centering 
     \begin{subfigure}{0.18\textwidth}
        \includegraphics[width=\linewidth]{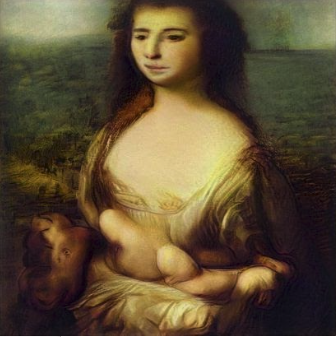}
        \caption*{towards emotion: \textbf{awe}}
      \end{subfigure}
    \begin{subfigure}{0.18\textwidth}
        \includegraphics[width=\linewidth]{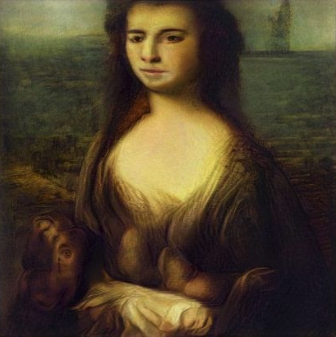}
        \caption*{}

    \end{subfigure}
    \begin{subfigure}{0.18\textwidth}
        \includegraphics[width=\linewidth]{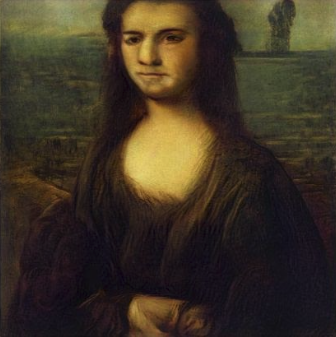}
        \caption*{Inversion on Mona Lisa}
    \end{subfigure}
    \begin{subfigure}{0.18\textwidth}
        \includegraphics[width=\linewidth]{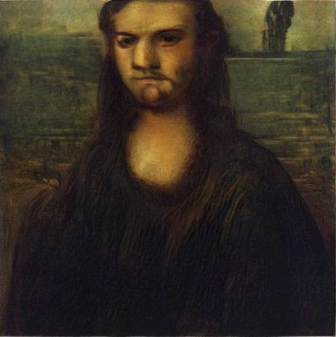}
        \caption*{}
    \end{subfigure}
    \begin{subfigure}{0.18\textwidth}
        \includegraphics[width=\linewidth]{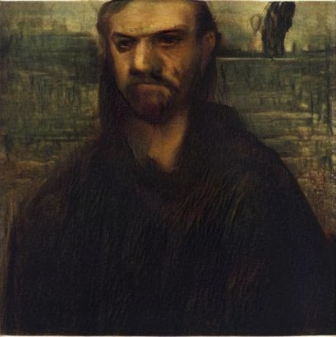}
        \caption*{towards emotion: \textbf{fear}}
    \end{subfigure}
    \captionsetup{width=.9\textwidth}
    \caption{The image at the center is the result of a GAN inversion process for the original \emph{La Gioconda} (Mona Lisa) painting. Then we apply condition-based vector arithmetic to the corresponding $\wvec$ vector: a transformation vector between the emotions \emph{awe} and \emph{fear}. Note that the network has acquired several biases: when we apply the transformation towards \emph{fear}, the portrayed woman turns into a grim-looking man. Towards \emph{awe}, the skin tone becomes noticeably lighter.}
    \label{fig:mona-lisa-arithmetic}
\end{figure*}

We use the following methodology to find $\vec{t}_{c_1,c_2}$: We sample $\wvec_{c_1}$ and $\wvec_{c_2}$ as described above with the same random noise vector $\zvec$ but different conditions and compute their difference. 
We repeat this process for a large number of randomly sampled $\zvec$. 
Then we compute the mean of the thus obtained differences, which serves as our transformation vector $\vec{t}_{c_1,c_2}$. 
As shown in \autoref{eq:w-vector}, this is equivalent to computing the difference between the conditional centers of mass of the respective conditions:

\begin{equation}
\begin{aligned}
  \vec{t}_{c_1,c_2} ={} & \mathbb{E}_{\zvec \sim P(\zvec)}[\fc(\zvec, c_2) - \fc(\zvec, c_1)]\\
  ={} & \mathbb{E}_{\zvec \sim P(\zvec)}[\fc(\zvec, c_2)] - \mathbb{E}_{\zvec \sim P(\zvec)}[\fc(\zvec, c_1)]\\
  ={} & \bar{\wvec}_{c_2} - \bar{\wvec}_{c_1}.
  \label{eq:w-vector}
 \end{aligned}
\end{equation}
Obviously, when we swap $c_1$ and $c_2$, the resulting transformation vector is negated:
\begin{equation}
     \vec{t}_{c_1,c_2} = -\vec{t}_{c_2,c_1}.
\end{equation}
Simple \textit{conditional interpolation} is the interpolation between two vectors in $W$ that were produced with the same $\zvec$ but different conditions. 
In contrast to conditional interpolation, our translation vector can be applied even to vectors in $W$ for which we do not know the corresponding $\zvec$ or condition. 
This is the case in GAN inversion, where the $\wvec$ vector corresponding to a real-world image is iteratively computed. 
One such example can be seen in \autoref{fig:mona-lisa-arithmetic}, where the GAN inversion process is applied to the original Mona Lisa painting. 
For the GAN inversion, we used the method proposed by Karras~\etal, which utilizes additive ramped-down noise~\cite{karras-stylegan2}.
To improve the low reconstruction quality, we optimized for the extended $W^+$ space and also optimized for the $P^+$ and improved $P_N^+$ space proposed by Zhu~\etal~\cite{zhu2021improved}.
We decided to use the reconstructed embedding from the $P^+$ space, as the resulting image was significantly better than the reconstructed image for the $W^+$ space and equal to the one from the $P_N^+$ space.
The resulting approximation of the Mona Lisa is clearly distinct from the original painting, which we attribute to the fact that human proportions in general are hard to learn for our network.

\section{Exploring Multi-Conditional StyleGAN}
\label{sec:multi-conditional}
With data for multiple conditions at our disposal, we of course want to be able to use all of them simultaneously to guide the image generation. 
This could be skin, hair, and eye color for faces, or art style, emotion, and painter for EnrichedArtEmis. 
Let $S$ be the set of unique conditions. 
We then define a multi-condition $\zeta$ as being comprised of multiple sub-conditions $c_s$, where $s \in S$. The available sub-conditions in EnrichedArtEmis are listed in \autoref{tab:labels-art-dataset}.

\subsection{Creating a Multi-Conditional Condition Vector}
\label{sec:mmdcv}
To use a multi-condition $\zeta$ during the training process for StyleGAN, we need to find a vector representation that can be fed into the network alongside the random noise vector. 
We do this by first finding a vector representation for each sub-condition $c_s$. 
Then we concatenate these individual representations. 
For EnrichedArtEmis, we have three different types of representations for sub-conditions. 
Emotions are encoded as a probability distribution vector with nine elements, which is the number of emotions in EnrichedArtEmis. 
Each element denotes the percentage of annotators that labeled the corresponding emotion. 
Categorical conditions such as painter, art style and genre are one-hot encoded. For textual conditions, such as content tags and explanations, we use a pretrained TinyBERT embedding~\cite{jiao2020tinybert}.

\begin{figure}[ht]
  \centering
  \begin{subfigure}{0.48\linewidth}
    \includegraphics[width=1\linewidth]{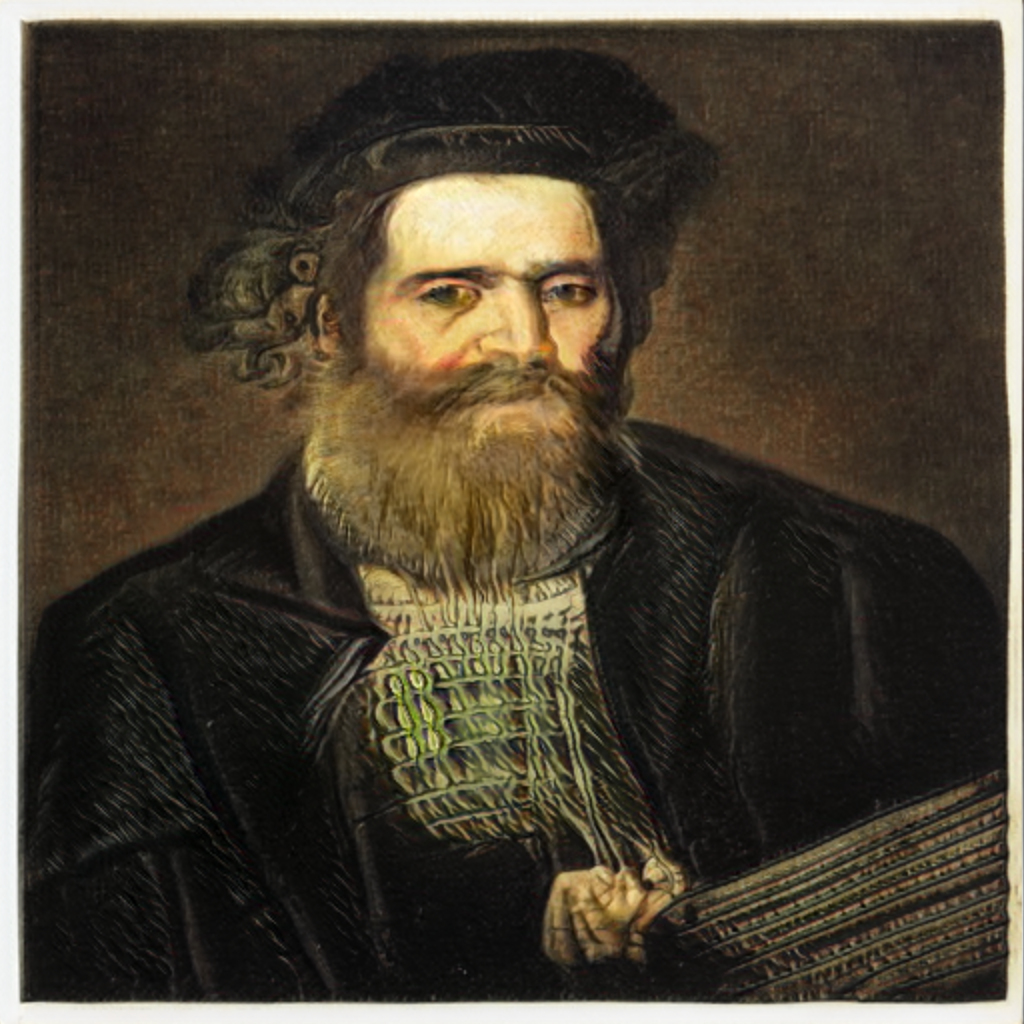}
    \caption*{Emotion: \textbf{anger}, Genre: \textbf{portrait}, Style: \textbf{Baroque}, Painter: \textbf{Rembrandt}, Content tag: \textbf{gentleman}}
  \end{subfigure}
  \hfill
  \begin{subfigure}{0.48\linewidth}
    \includegraphics[width=1\linewidth]{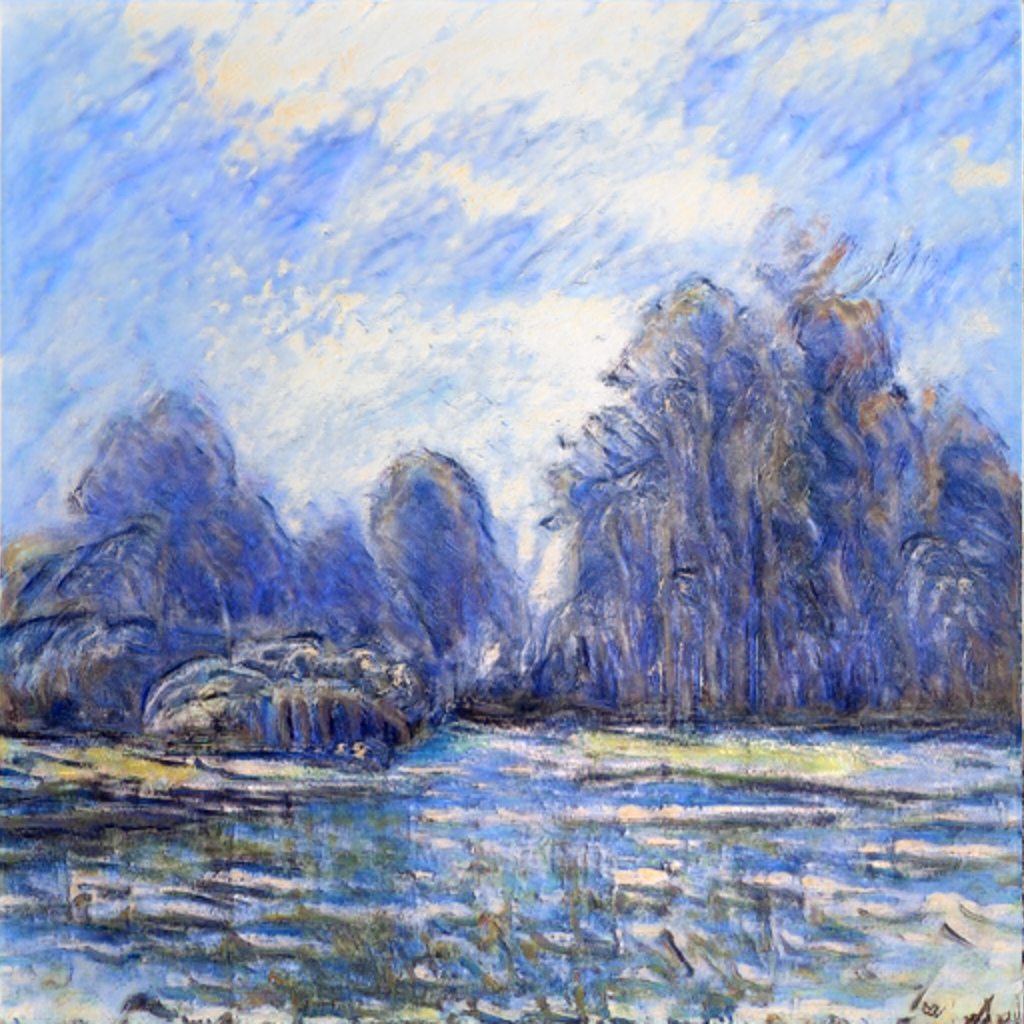}
    \caption*{Emotion: \textbf{awe}, Genre: \textbf{landscape}, Style: \textbf{Impressionism}, Painter: \textbf{Monet}, Content tag: \textbf{trees}}
  \end{subfigure}
  \caption{Paintings produced by a multi-conditional StyleGAN model trained with the conditions \emph{emotion}, \emph{genre}, \emph{style}, \emph{painter} and \emph{content} tags. More details about the conditions are given in \autoref{tab:labels-art-dataset}.}
  \label{fig:multicond-examples}
\end{figure}

\begin{figure}[ht]
  \centering
  \begin{subfigure}{0.32\linewidth}
    \includegraphics[width=1\linewidth]{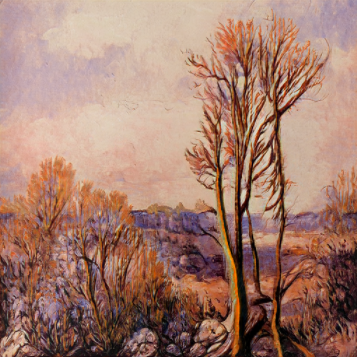}
    \caption*{orange}
  \end{subfigure}
  \hfill
\begin{subfigure}{0.32\linewidth}
    \includegraphics[width=1\linewidth]{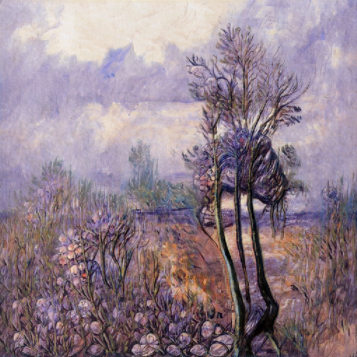}
    \caption*{violet}
  \end{subfigure}
  \hfill
  \begin{subfigure}{0.32\linewidth}
    \includegraphics[width=1\linewidth]{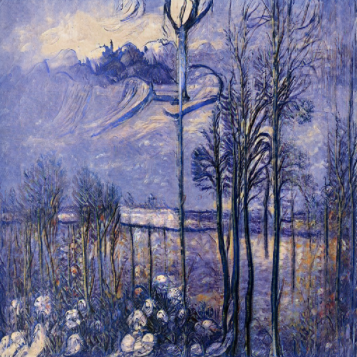}
    \caption*{blue}
  \end{subfigure}
  \caption{Paintings produced by a multi-conditional StyleGAN model with conditions \emph{emotion}: \textbf{contentment}, \emph{genre}: \textbf{landscape}, \emph{style}: \textbf{Impressionism}, \emph{painter}: \textbf{Monet}, and varying colors as a \emph{content} tag. All images are generated with the same random noise vector $\zvec$.}
  \label{fig:color}
\end{figure}

\begin{figure}[ht]
\captionsetup[subfigure]{justification=centering}
  \centering
  \begin{subfigure}{0.48\linewidth}
    \includegraphics[width=1\linewidth]{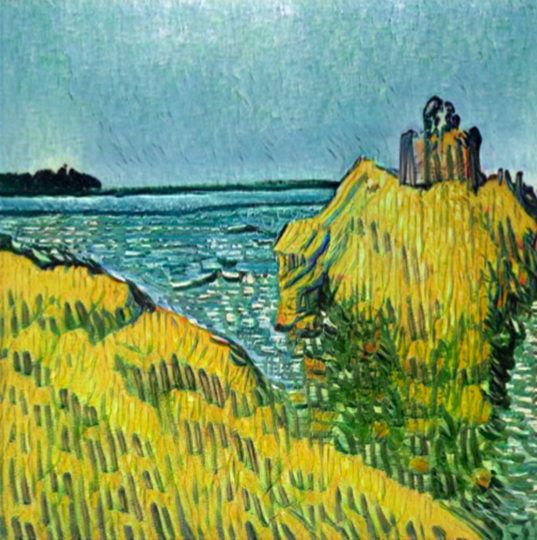}
    \caption*{Style: \textbf{Post-Impressionism},\\ Painter: \textbf{Vincent van Gogh}}
  \end{subfigure}
  \hfill
  \begin{subfigure}{0.48\linewidth}
    \includegraphics[width=1\linewidth]{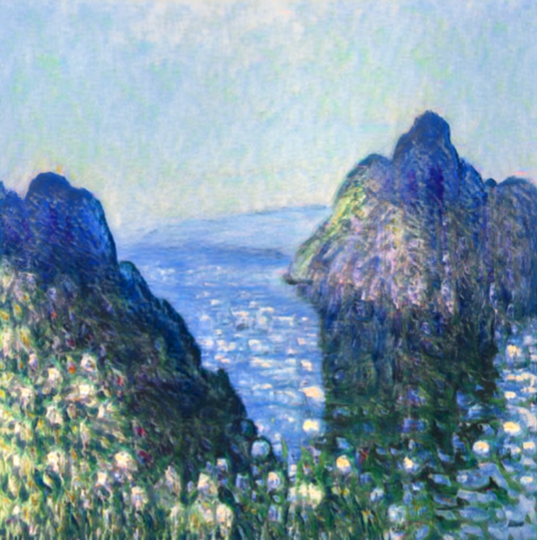}
    \caption*{Style: \textbf{Impressionism},\\ Painter: \textbf{Claude Monet}}
  \end{subfigure}
  \caption{Comparison of paintings produced by a multi-conditional StyleGAN model for the painters \emph{Monet} and \emph{van Gogh}. Both paintings are produced using the same random noise $\zvec$ and with \emph{genre}: \textbf{landscape}, \emph{emotion}: \textbf{contentment}, \emph{content} tag: \textbf{water}.
}
  \label{fig:gogh-vs-monet}
\end{figure}

With this setup, multi-conditional training and image generation with StyleGAN is possible. 
In \autoref{fig:multicond-examples} and \autoref{fig:color}, we can see paintings produced by this multi-conditional generation process. 
It is worth noting that some conditions are more subjective than others.
The emotions a painting evoke in a viewer are highly subjective and may even vary depending on external factors such as mood or stress level. 
Nevertheless, we observe that most sub-conditions are reflected rather well in the samples.
In \autoref{fig:gogh-vs-monet}, we compare our network's renditions of Vincent van Gogh and Claude Monet. For van Gogh specifically, the network has learned to imitate the artist's famous brush strokes and use of bold colors. However, while these samples might depict good imitations, they would by no means fool an art expert.

\subsection{Wildcard Generation}
A multi-conditional StyleGAN model allows us to exert a high degree of influence over the generated samples. 
For example, when using a model trained on the sub-conditions \emph{emotion}, \emph{art style}, \emph{painter}, \emph{genre}, and \emph{content} tags, we can attempt to generate ``awe-inspiring, impressionistic landscape paintings with trees by Monet". 
However, this degree of influence can also become a burden, as we \textit{always} have to specify a value for every sub-condition that the model was trained on. 
Considering real-world use cases of GANs, such as stock image generation, this is an undesirable characteristic, as users likely only care about a select subset of the entire range of conditions. 
For instance, a user wishing to generate a stock image of a smiling businesswoman may not care specifically about eye, hair, or skin color.

Therefore, we propose \textit{wildcard generation}: For a multi-condition $\zeta$, we wish to be able to replace arbitrary sub-conditions $c_s$ with a \textit{wildcard mask} and still obtain samples that adhere to the parts of $\zeta$ that were not replaced. 
The model has to interpret this wildcard mask in a meaningful way in order to produce sensible samples.
As our wildcard mask, we choose replacement by a zero-vector.
Specifically, any sub-condition $c_s$ within $\zeta$ that is not specified is replaced by a zero-vector of the same length. 

To ensure that the model is able to handle such $\zeta$, we also integrate this into the training process with a \textit{stochastic condition masking} regime. 
Whenever a sample is drawn from the dataset, $k$ sub-conditions are randomly chosen from the entire set of sub-conditions. 
Then, each of the chosen sub-conditions is masked by a zero-vector with a probability $p$. 
We choose this way of selecting the masked sub-conditions in order to have two hyper-parameters $k$ and $p$. 
If $k$ is too close to the number of available sub-conditions, the training process collapses because the generator receives too little information as too many of the sub-conditions are masked. 
If $k$ is too low, the generator might not learn to generalize towards cases where more conditions are left unspecified.
The probability $p$ can be used to adjust the effect that the stochastic conditional masking effect has on the entire training process.

\begin{figure*}[!htbp]
  \centering
  \begin{subfigure}{0.24\textwidth}
    \includegraphics[width=1\linewidth]{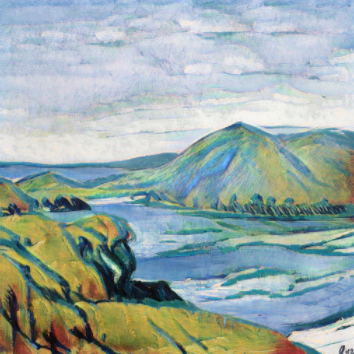}
  \end{subfigure}
  \hfill
\begin{subfigure}{0.24\textwidth}
    \includegraphics[width=1\linewidth]{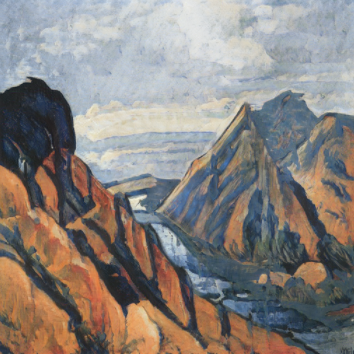}
  \end{subfigure}
  \hfill
  \begin{subfigure}{0.24\textwidth}
    \includegraphics[width=1\linewidth]{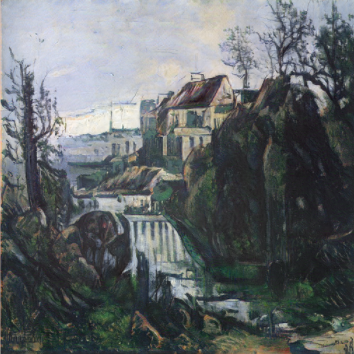}
  \end{subfigure}
  \hfill
  \begin{subfigure}{0.24\textwidth}
    \includegraphics[width=1\linewidth]{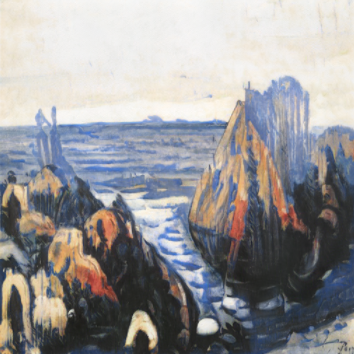}
  \end{subfigure}
  \caption{Paintings produced by a multi-conditional StyleGAN model with the conditions \emph{genre}: \textbf{landscape}, \emph{content} tag: \textbf{mountains}, and \emph{style}, \emph{painter}, and \emph{emotion} replaced by a \textit{wildcard} zero-vector.}
  \label{fig:wildcard-mountains}
\end{figure*}

In \autoref{fig:wildcard-mountains}, we can see the result of such a wildcard generation. 
The paintings match the specified condition of ``landscape painting with mountains". 
It is worth noting however that there is a degree of \textit{structural similarity} between the samples. 
While the samples are still visually distinct, we observe similar subject matter depicted in the same places across all of them. 
For example, the lower left corner as well as the center of the right third are occupied by mountainous structures.
This seems to be a weakness of wildcard generation when specifying few conditions as well as our multi-conditional StyleGAN in general, especially for rare combinations of sub-conditions.

\section{Evaluation}

\newcommand\GANt{\emph{GAN}$_{\text{\textsc{T}}}$}
\newcommand\GANesg{\emph{GAN}$_{\textsc{ESG}}$}
\newcommand\GANesgpt{\emph{GAN}$_{\textsc{ESGPT}}$}

The key characteristics that we seek to evaluate are the 
quality of the generated images and to what extent they adhere to the provided conditions.
Although we meet the main requirements proposed by Baluja~\etal to produce pleasing computer-generated images~\cite{baluja94}, the question remains whether our generated artworks are of sufficiently high quality.

\paragraph{Data.}
All models are trained on the EnrichedArtEmis dataset described in Section~\ref{sec:dataset}, using a standardized 512$\times$512 resolution obtained via resizing and optional cropping.
In total, we have two conditions (\textit{emotion} and \textit{content tag}) that have been evaluated by non art experts and three conditions (\textit{genre}, \textit{style}, and \textit{painter}) derived from meta-information.

\paragraph{Models.} One of our GANs has been exclusively trained using the \textit{content tag} condition of each artwork, which we denote as \GANt{}.
The remaining GANs are multi-conditioned:
The second \GANesg{} is trained on \textit{emotion}, \textit{style}, and \textit{genre}, whereas the third \GANesgpt{} includes the conditions of both \GANt{} and \GANesg{} in addition to the condition \textit{painter}. All GANs are trained with default parameters and an output resolution of 512$\times$512.

\subsection{Evaluation Metrics}

In the literature on GANs, a number of metrics have been found to correlate with the image quality
and hence have gained widespread adoption \cite{szegedy2015rethinking,devries19,binkowski21}. Additionally, we also conduct a manual qualitative analysis.

\subsubsection{Qualitative Evaluation}
Our first evaluation is a qualitative one considering to what extent the models are able to consider the specified conditions, based on a manual assessment.
For this, we first define the function $b(i, c)$ to capture whether an image matches its specified condition after manual evaluation as a numerical value:
\begin{equation*}
b(i, c) = \begin{cases}
      1 & \text{image } i \text{ matches the condition } c \\
      0 & \text{else}
   \end{cases}
   \label{eq:b:qual}
\end{equation*}
Given a sample set $S$, where each entry $s \in S$ consists of the image $s_\mathrm{img}$ and the condition vector $s_{c}$, we summarize the overall correctness as $e_\mathrm{qual}(S)$, defined as follows.
\begin{equation}
    \begin{aligned}
    e_\mathrm{qual}(s) = {} & \dfrac{1}{d}\sum_{i=1}^{d} b(s_\mathrm{img},s_{c_{i}})\\
    e_\mathrm{qual}(S) = {} & \dfrac{1}{|S|}\sum_{s \in S} e_\mathrm{qual}(s)
      \label{eq:e_qual}
    \end{aligned}
\end{equation}

We determine a suitable sample sizes $n_\mathrm{qual}$ for $S$ based on the condition shape vector $\vec{c}_\mathrm{shape}=\left[c_1, \dots, c_{d}\right] \in \mathbb{R}^{d}$ for a given GAN. This vector of dimensionality $d$ captures the number of condition entries for each condition, e.g., $[9, 30, 31]$ for \GANesg. 
Here, we have a tradeoff between significance and feasibility.
While one traditional study suggested 10\% of the given combinations \cite{bohanec92}, this quickly becomes impractical when considering highly multi-conditional models as in our work.
Thus, for practical reasons, $n_\mathrm{qual}$ is capped at a threshold of $n_{\max}=100$:
\begin{equation}
  n_\mathrm{qual} = \min\left( \left\lceil \dfrac{1}{10}\, \prod_{i=1}^{d} c_i \right\rceil + 10, n_{\max}\right)
  \label{eq:n_qual}
\end{equation}

The proposed method enables us to assess how well different GANs are able to match the desired conditions.
The results of our GANs are given in \autoref{tab:q_evaluation}.

\begin{table}[ht]
\centering
\begin{tabular}{lccc}
\toprule
\textbf{Metric}   & \textbf{\GANt} &  \textbf{\GANesg} & \textbf{\GANesgpt}
\\ \midrule 
$n_\mathrm{qual}$    & 77  & 100 & 100 \\
$e_\mathrm{qual}$    & 0.91 & 0.88 & 0.83 \\
\bottomrule
\end{tabular}
\caption{Qualitative evaluation for the (multi-)conditional GANs.}
\label{tab:q_evaluation}
\end{table}

\subsubsection{Fréchet Inception Distance (FID)}
Apart from using classifiers or Inception Scores (IS)~\cite{salimans16}, which focus on diversity, the Fréchet Inception Distance (FID) is among the most widely adopted metrics to evaluate GANs~\cite{heusel2018gans}.
The FID estimates the quality of a collection of generated images by using the embedding space of the pretrained InceptionV3 model~\cite{szegedy2015rethinking}.
In this case, the term ``quality'' refers to the proximity of the real and generated data, specifically the proximity of the multivariate Gaussian distributions of the real and generated data.
The characteristic difference is the embedding function $f$ that embeds an image tensor into a learned feature space.
The variable $\vec{x}$ within the embedding function $f$ corresponds to the image vector.
For the FID, we define $\Sigma$ and $\muvec$ of the Fréchet distance (\autoref{eq:fd}) as follows: 
\begin{equation}
    \begin{aligned}
    \muvec = {} & \dfrac{1}{N}\sum_{i=0}^N f(\vec{x}^{(i)}), \\
    \Sigma = {} & \dfrac{1}{N-1} \sum_{i=0}^N (f(\vec{x}^{(i)})-\muvec)(f(\vec{x}^{(i)})-\muvec)^\intercal
     \label{eq:fid}
    \end{aligned}
\end{equation}
Generally speaking, a lower score represents a closer proximity to the original dataset.
A score of 0 on the other hand corresponds to exact copies of the real data.
In order to reliably calculate the FID score, a sample size of 50,000 images is recommended~\cite{szegedy2015rethinking}.

\begin{figure}
  \centering
    \includegraphics[width=1\linewidth]{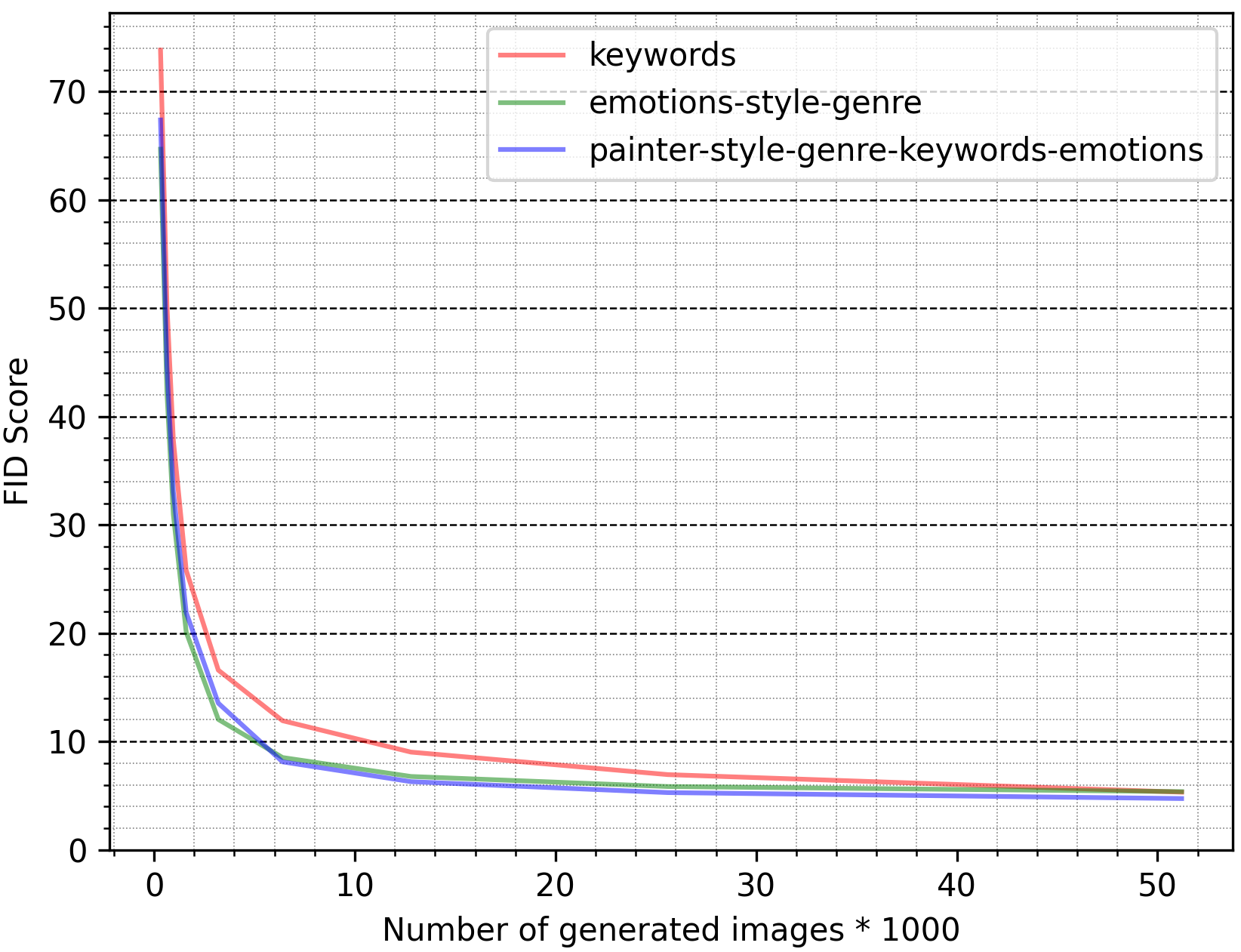}
    \caption{FID Convergence for different GAN models.}
    \label{fig:fid_k}
\end{figure}

The results in \autoref{fig:fid_k} highlight the increased volatility at a low sample size and their convergence to their true value for the three different GAN models.

\begin{figure}
  \centering
    \includegraphics[width=1\linewidth]{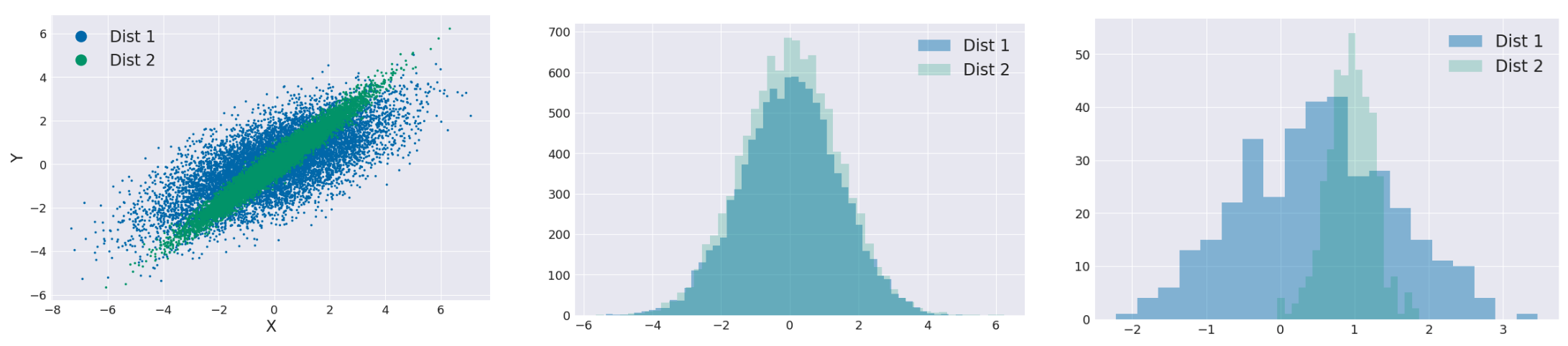}
    \caption{Left: samples from two multivariate Gaussian distributions. Center: Histograms of marginal distributions for Y. Right: Histogram of conditional distributions for Y. Images from DeVries~\etal~\cite{devries19}.}
    \label{fig:fjd_distibution}
\end{figure}

Unfortunately, most of the metrics used to evaluate GANs focus on measuring the similarity between generated and real images without addressing whether conditions are met appropriately~\cite{devries19}.
The FID, in particular, only considers the marginal distribution of the output images and therefore does not include any information regarding the conditioning.
This means that our networks may be able to produce closely related images to our original dataset without any regard for conditions and still obtain a good FID score.
\autoref{fig:fjd_distibution} illustrates the differences of two multivariate Gaussian distributions mapped to the marginal and the conditional distributions.
Due to the downside of not considering the conditional distribution for its calculation, 
we cannot use the FID score to evaluate how good the conditioning of our GAN models are.

\subsubsection{Fréchet Joint Distance (FJD)}
Accounting for both conditions and the output data is possible with the Fréchet Joint Distance (FJD) by DeVries~\etal.~\cite{devries19}.
It involves calculating the Fréchet Distance (\autoref{eq:fd}) over the joint image--conditioning embedding space.
We can achieve this using a merging function
\begin{equation}
     g = \left[f(\vec{x}^{(i)}),\, \alpha h(\vec{y}^{(i)})\right]
 \label{eq:fjd_g}
\end{equation}
that concatenates representations for the image vector $\vec{x}$ and the conditional embedding $\vec{y}$. The representation for the latter is obtained using an embedding function $h$ that embeds our multi-conditions as stated in Section~\ref{sec:mmdcv}.
A scaling factor $\alpha$ allows us to flexibly adjust the impact of the conditioning embedding compared to the vanilla FID score. Setting $\alpha = 0$ corresponds to the evaluation of the marginal distribution of the FID.

The above merging function $g$ replaces the original invocation of $f$ in the FID computation to evaluate the conditional distribution of the data.

By calculating the FJD, we have a metric that simultaneously compares the image quality, conditional consistency, and intra-condition diversity.
The main downside is the comparability of GAN models with different conditions.
DeVries~\etal.~\cite{devries19} mention the importance of maintaining the same embedding function, reference distribution, and $\alpha$ value for reproducibility and consistency.
Simply adjusting $\alpha$ for our GAN models to balance changes does not work for our GAN models, due to the varying sizes of the individual sub-conditions and their structural differences.
We can also tackle this compatibility issue by addressing every condition of a GAN model individually.

\subsubsection{Intra-Fréchet Inception Distance (I-FID)}
Our implementation of Intra-Fréchet Inception Distance (I-FID) is inspired by Takeru~\etal~\cite{takeru18} and allows us to compare the impact of the individual conditions.
Additionally, the I-FID still takes image quality, conditional consistency, and intra-class diversity into account.

However, this approach scales poorly with a high number of unique conditions and a small sample size such as for our \GANesgpt. Also, the computationally intensive FID calculation must be repeated for each condition, and because FID behaves poorly when the sample size is small~\cite{binkowski21}.
We resolve this issue by only selecting 50\% of the condition entries $ce$ within the corresponding distribution.
Therefore, we select the $ce$ of each condition by size in descending order until we reach the given threshold.
We have found that 50\% is a good estimate for the I-FID score and closely matches the accuracy of the complete I-FID.
Hence, we can reduce the computationally exhaustive task of calculating the I-FID for all the outliers.
After determining the set of $ce$ for every condition, we collect a set of condition samples $S$ from the dataset matching $ce$ and repeat this procedure for every condition.
Then, we generate the intra-conditional images based on $S$ and calculate its local FID score.
This procedure is repeated for every $ce$.
Afterwards, the average of each condition is calculated.
Lastly, we also compute the condition average that represents our I-FID score.

\subsubsection{Hybrid Evaluation Metric}
Due to the large variety of conditions and the ongoing problem of recognizing objects or characteristics in general in artworks~\cite{cai15}, we further propose a combination of qualitative and quantitative evaluation scoring for our GAN models, inspired by Bohanec~\etal~\cite{bohanec92}.
For this, we first compute the quantitative metrics as well as the qualitative score given earlier by \autoref{eq:e_qual}.
Subsequently, 
we compute a weighted average: 
\begin{equation}
    \begin{aligned}
   e_\mathrm{art} = {} & \frac{e_\text{I-FID} + e_\text{FJD}}{2}\, (2-e_\mathrm{qual})
     \label{eq:e_art}
 \end{aligned}
\end{equation}
Hence, we can compare our multi-conditional GANs in terms of image quality, conditional consistency, and intra-conditioning diversity. 

\subsection{Results}

\begin{table}
\begin{tabular}{@{}lrrr@{}}
\toprule
\textbf{Metric} & \textbf{\GANt} & \textbf{\GANesg} & \textbf{\GANesgpt}\\
\midrule 
FID                     & 5.38  & 5.37  & 4.67\\
Emotion Intra-FID       & --    & 10.51 & 9.74\\ 
Style Intra-FID         & --    & 9.23  & 7.98\\ 
Genre Intra-FID         & --    & 8.19  & 7.31\\ 
Painter Intra-FID       & --    & --    & 8.65\\ 
Content-Tag Intra-FID   & 5.46  & --    & 6.83\\
Intra-FID (average)     & 5.46  & 9.31  & 8.10\\ 
FJD ($\alpha = 0.5$)    & 9.42  & 9.29  & 8.47\\
Qualitative results ($e_\mathrm{qual}$) & 0.91  & 0.88  & 0.83\\
Hybrid metric ($e_\mathrm{art}$)               & 8.11  & 10.42 & 9.69\\
\bottomrule
\end{tabular}
\caption{Overall evaluation using quantitative metrics as well as our proposed hybrid metric for our (multi-)conditional GANs.}
\label{tab:evaluation}
\end{table}

\paragraph{Overview.}
The results are given in \autoref{tab:evaluation}. Given a particular GAN model, we followed previous work \cite{szegedy2015rethinking} and generated at least 50,000 multi-conditional artworks for each quantitative experiment in the evaluation. The results reveal that the quantitative metrics mostly match the actual results of manually checking the presence of every condition.
However, with an increased number of conditions, the qualitative results start to diverge from the quantitative metrics.
This validates our assumption that the quantitative metrics do not perfectly represent our perception when it comes to the evaluation of multi-conditional images.
Despite the small sample size, we can conclude that our manual labeling of each condition acts as an uncertainty score for the reliability of the quantitative measurements. 
We conjecture that the worse results for \GANesgpt{} may be caused by outliers, due to the higher probability of producing rare condition combinations.

Overall, we find that we do not need an additional classifier that would require large amounts of training data to enable a reasonably accurate assessment.
Instead, we can use our $e_\mathrm{art}$ metric from \autoref{eq:e_art} to put the considered GAN evaluation metrics in context.
All in all, somewhat unsurprisingly, the conditional \GANt{} produces more accurate results in comparison to multi-conditional GANs with many different conditions. However, the latter provide substantial control that can be very useful when generating art.

\begin{figure}
  \centering
  \begin{subfigure}{0.48\linewidth}
    \includegraphics[width=1\linewidth]{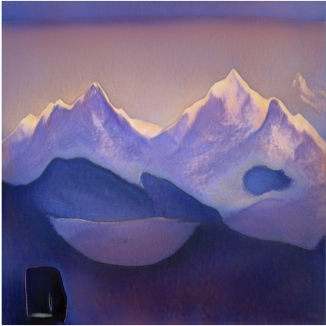}
    \caption*{Generated artwork.}
  \end{subfigure}
  \hfill
  \begin{subfigure}{0.48\linewidth}
    \includegraphics[width=1\linewidth]{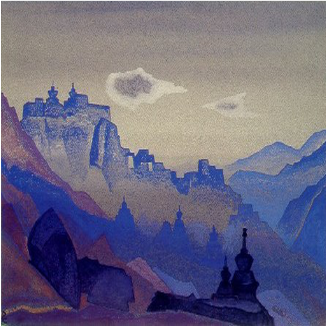}
    \caption*{Nearest neighbor in dataset.}
  \end{subfigure}
  \caption{Generated artwork and its nearest neighbor in the training data based on a \textit{perceptual similarity} metric~\cite{zhang2018perceptual}.}
  \label{fig:nn}
\end{figure}

\paragraph{Analysis.}
In order to eliminate the possibility that a model is merely replicating images from the training data, we compare a generated image to its nearest neighbors in the training data.
To find these nearest neighbors, we use a \textit{perceptual similarity} measure~\cite{zhang2018perceptual}, which measures the similarity of two images embedded in a deep neural networks' intermediate feature space.
Using this method, we did not find any generated image to be a near-identical copy of an image in the training dataset. A typical example of a generated image and its nearest neighbor in the training dataset is given in \autoref{fig:nn}. 

\paragraph{Discussion.}
In the literature on GANs, a number of quantitative metrics have been found to correlate with the image quality
and hence have gained widespread adoption \cite{szegedy2015rethinking, devries19, binkowski21}. 
The proposed methods do not explicitly judge the visual quality of an image but rather focus on how well the images produced by a GAN match those in the original dataset, both generally and with regard to particular conditions.
Hence, the \emph{image quality} here is considered with respect to a particular dataset and model.
We consider the definition of creativity of Dorin and Korb, which evaluates the probability to produce certain representations of patterns~\cite{dorin09} and extend it to the GAN architecture.

Of course, historically, art has been evaluated qualitatively by humans. Such assessments, however, may be costly to procure and are also a matter of taste and thus it is not possible to obtain a completely objective evaluation. Our initial attempt to assess the quality was to train an InceptionV3 image classifier~\cite{szegedy2015rethinking} on subjective art ratings of the WikiArt dataset~\cite{mohammed2018artemo}. Such a rating may vary from 3 (like a lot) to -3 (dislike a lot), representing the average score of non art experts. However, this approach did not yield satisfactory results, as the classifier made seemingly arbitrary predictions.
We believe that this is due to the small size of the annotated training data (just 4,105 samples) as well as the inherent subjectivity and the resulting inconsistency of the annotations. Still, in future work, we believe that a broader qualitative evaluation by art experts as well as non-experts would be a valuable addition to our presented techniques.

\section{Conclusion}
In this paper, we have applied the powerful StyleGAN architecture to a large art dataset and investigated techniques to enable multi-conditional control.
The images that this trained network is able to produce are convincing and in many cases appear to be able to pass as human-created art.
Due to the nature of GANs, the created images of course may perhaps be viewed as imitations rather than as truly novel or creative art.
This stems from the objective function that is optimized during training, which encourages the model to imitate the training distribution as closely as possible. 
Our evaluation shows that automated quantitative metrics start diverging from human quality assessment as the number of conditions increases, especially due to the uncertainty of precisely classifying a condition.
To alleviate this challenge, we also conduct a qualitative evaluation and propose a hybrid score.

We further examined the conditional embedding space of StyleGAN and were able to learn about the conditions themselves.
Our proposed conditional truncation trick (as well as the conventional truncation trick) may be used to emulate specific aspects of creativity: novelty or \textit{unexpectedness}. 
To improve the fidelity of images to the training distribution at the cost of diversity, we propose interpolating \textit{towards} a (conditional) center of mass. 
However, in future work, we could also explore interpolating \textit{away} from it, thus increasing diversity and decreasing fidelity, i.e., increasing unexpectedness.
Furthermore, art is more than just the painting --- it also encompasses the story and events around an artwork.
A network such as ours could be used by a creative human to tell such a story; as we have demonstrated, condition-based vector arithmetic might be used to generate a series of connected paintings with conditions chosen to match a narrative.

As it stands, we believe creativity is still a domain where humans reign supreme. However, our work shows that humans may use artificial intelligence as a means of expressing or enhancing their creative potential.

{\small
\bibliographystyle{ieee_fullname}
\bibliography{egbib}

\begin{thebibliography}{10}\itemsep=-1pt

\bibitem{abdal2019image2stylegan}
Rameen Abdal, Yipeng Qin, and Peter Wonka.
\newblock {Image2StyleGAN}: {H}ow to embed images into the {StyleGAN} latent
  space?, 2019.

\bibitem{abdal2020image2stylegan}
Rameen Abdal, Yipeng Qin, and Peter Wonka.
\newblock {Image2StyleGAN++}: {H}ow to edit the embedded images?, 2020.

\bibitem{abdal2020styleflow}
Rameen Abdal, Peihao Zhu, Niloy Mitra, and Peter Wonka.
\newblock {StyleFlow}: {A}ttribute-conditioned exploration of
  {StyleGAN}-generated images using conditional continuous normalizing flows,
  2020.

\bibitem{achlioptas2021artemis}
Panos Achlioptas, Maks Ovsjanikov, Kilichbek Haydarov, Mohamed Elhoseiny, and
  Leonidas Guibas.
\newblock {ArtEmis}: {A}ffective language for visual art.
\newblock {\em CoRR}, abs/2101.07396, 2021.

\bibitem{arjovsky2017wasserstein}
Martin Arjovsky, Soumith Chintala, and Léon Bottou.
\newblock Wasserstein {GAN}, 2017.

\bibitem{baluja94}
Shumeet Baluja, Dean Pomerleau, and Todd Jochem.
\newblock Towards automated artificial evolution for computer-generated images.
\newblock {\em Connection Science}, 6(2-3):325--354, 1994.

\bibitem{binkowski21}
Mikolaj Binkowski, Danica~J. Sutherland, Michael Arbel, and Arthur Gretton.
\newblock Demystifying {MMD} {GAN}s, 2021.

\bibitem{bohanec92}
Marko Bohanec, Bozo Urh, and Vladislav Rajkovič.
\newblock Evaluating options by combined qualitative and quantitative methods.
\newblock {\em Acta Psychologica}, 80(1):67--89, 1992.

\bibitem{brock2018largescalegan}
Andrew Brock, Jeff Donahue, and Karen Simonyan.
\newblock Large scale {GAN} training for high fidelity natural image synthesis.
\newblock {\em CoRR}, abs/1809.11096, 2018.

\bibitem{cai15}
Hongping Cai, Qi Wu, Tadeo Corradi, and Peter Hall.
\newblock The cross-depiction problem: {C}omputer vision algorithms for
  recognising objects in artwork and in photographs, 2015.

\bibitem{devries2017modulating}
Harm de Vries, Florian Strub, Jérémie Mary, Hugo Larochelle, Olivier
  Pietquin, and Aaron Courville.
\newblock Modulating early visual processing by language, 2017.

\bibitem{devries19}
Terrance DeVries, Adriana Romero, Luis Pineda, Graham Taylor, and Michal
  Drozdzal.
\newblock On the evaluation of conditional {GANs}, 07 2019.

\bibitem{dorin09}
Alan Dorin and Kevin~B Korb.
\newblock Improbable creativity.
\newblock In Margaret Boden, Mark D'Inverno, and Jon McCormack, editors, {\em
  Computational Creativity: {A}n Interdisciplinary Approach}, number 09291 in
  Dagstuhl Seminar Proceedings, Dagstuhl, Germany, 2009. Schloss Dagstuhl -
  Leibniz-Zentrum fuer Informatik, Germany.

\bibitem{dowson1982frechet}
D.C Dowson and B.V Landau.
\newblock The {Fréchet} distance between multivariate normal distributions.
\newblock {\em Journal of Multivariate Analysis}, 12(3):450--455, 1982.

\bibitem{elgammal2017can}
Ahmed Elgammal, Bingchen Liu, Mohamed Elhoseiny, and Marian Mazzone.
\newblock {CAN}: Creative adversarial networks, generating "art" by learning
  about styles and deviating from style norms, 2017.

\bibitem{goodfellow2014generative}
Ian~J. Goodfellow, Jean Pouget-Abadie, Mehdi Mirza, Bing Xu, David
  Warde-Farley, Sherjil Ozair, Aaron Courville, and Yoshua Bengio.
\newblock {Generative Adversarial Networks}, 2014.

\bibitem{heusel2018gans}
Martin Heusel, Hubert Ramsauer, Thomas Unterthiner, Bernhard Nessler,
  G{\"{u}}nter Klambauer, and Sepp Hochreiter.
\newblock {GANs} trained by a two time-scale update rule converge to a {Nash}
  equilibrium.
\newblock {\em CoRR}, abs/1706.08500, 2017.

\bibitem{jiao2020tinybert}
Xiaoqi Jiao, Yichun Yin, Lifeng Shang, Xin Jiang, Xiao Chen, Linlin Li, Fang
  Wang, and Qun Liu.
\newblock {TinyBERT}: {D}istilling {BERT} for natural language understanding,
  2020.

\bibitem{karras2018progressive}
Tero Karras, Timo Aila, Samuli Laine, and Jaakko Lehtinen.
\newblock Progressive growing of {GANs} for improved quality, stability, and
  variation, 2018.

\bibitem{karras-stylegan2-ada}
Tero Karras, Miika Aittala, Janne Hellsten, Samuli Laine, Jaakko Lehtinen, and
  Timo Aila.
\newblock Training {Generative Adversarial Networks} with limited data.
\newblock {\em CoRR}, abs/2006.06676, 2020.

\bibitem{karras2020training}
Tero Karras, Miika Aittala, Janne Hellsten, Samuli Laine, Jaakko Lehtinen, and
  Timo Aila.
\newblock Training {Generative Adversarial Networks} with limited data, 2020.

\bibitem{karras2019stylebased}
Tero Karras, Samuli Laine, and Timo Aila.
\newblock A style-based generator architecture for {Generative Adversarial
  Networks}, 2019.

\bibitem{karras-stylegan2}
Tero Karras, Samuli Laine, Miika Aittala, Janne Hellsten, Jaakko Lehtinen, and
  Timo Aila.
\newblock Analyzing and improving the image quality of {StyleGAN}.
\newblock {\em CoRR}, abs/1912.04958, 2019.

\bibitem{karras2020analyzing}
Tero Karras, Samuli Laine, Miika Aittala, Janne Hellsten, Jaakko Lehtinen, and
  Timo Aila.
\newblock Analyzing and improving the image quality of {StyleGAN}, 2020.

\bibitem{liu2020sketchtoart}
Bingchen Liu, Kunpeng Song, and Ahmed Elgammal.
\newblock {Sketch-to-Art}: {S}ynthesizing stylized art images from sketches,
  2020.

\bibitem{mccormack2019autonomy}
Jon McCormack, Toby Gifford, and Patrick Hutchings.
\newblock Autonomy, authenticity, authorship and intention in computer
  generated art, 2019.

\bibitem{mirza2014conditional}
Mehdi Mirza and Simon Osindero.
\newblock Conditional {Generative Adversarial Nets}, 2014.

\bibitem{miyato2018cgans}
Takeru Miyato and Masanori Koyama.
\newblock {cGANs} with projection discriminator, 2018.

\bibitem{takeru18}
Takeru Miyato and Masanori Koyama.
\newblock {cGANs} with projection discriminator.
\newblock {\em CoRR}, abs/1802.05637, 2018.

\bibitem{mohammed2018artemo}
Saif~M. Mohammad and Svetlana Kiritchenko.
\newblock An annotated dataset of emotions evoked by art.
\newblock In {\em Proceedings of the 11th Edition of the Language Resources and
  Evaluation Conference (LREC-2018)}, Miyazaki, Japan, 2018.

\bibitem{nitzan2020face}
Yotam Nitzan, Amit Bermano, Yangyan Li, and Daniel Cohen-Or.
\newblock Face identity disentanglement via latent space mapping, 2020.

\bibitem{odena2017conditional}
Augustus Odena, Christopher Olah, and Jonathon Shlens.
\newblock Conditional image synthesis with auxiliary classifier {GANs}, 2017.

\bibitem{pan2020exploiting}
Xingang Pan, Xiaohang Zhan, Bo Dai, Dahua Lin, Chen~Change Loy, and Ping Luo.
\newblock Exploiting deep generative prior for versatile image restoration and
  manipulation, 2020.

\bibitem{park2018mcgan}
Hyojin Park, YoungJoon Yoo, and Nojun Kwak.
\newblock {MC-GAN}: {M}ulti-conditional {Generative Adversarial Network} for
  image synthesis, 2018.

\bibitem{radford2016unsupervised}
Alec Radford, Luke Metz, and Soumith Chintala.
\newblock Unsupervised representation learning with deep convolutional
  {Generative Adversarial Networks}, 2016.

\bibitem{salimans16}
Tim Salimans, Ian~J. Goodfellow, Wojciech Zaremba, Vicki Cheung, Alec Radford,
  and Xi Chen.
\newblock Improved techniques for training {GANs}.
\newblock {\em CoRR}, abs/1606.03498, 2016.

\bibitem{shen2020interpreting}
Yujun Shen, Jinjin Gu, Xiaoou Tang, and Bolei Zhou.
\newblock Interpreting the latent space of {GANs} for semantic face editing,
  2020.

\bibitem{szegedy2015rethinking}
Christian Szegedy, Vincent Vanhoucke, Sergey Ioffe, Jonathon Shlens, and
  Zbigniew Wojna.
\newblock Rethinking the {Inception} architecture for computer vision, 2015.

\bibitem{Ulyanov_2020}
Dmitry Ulyanov, Andrea Vedaldi, and Victor Lempitsky.
\newblock Deep image prior.
\newblock {\em International Journal of Computer Vision}, 128(7):1867–1888,
  Mar 2020.

\bibitem{voynov2020unsupervised}
Andrey Voynov and Artem Babenko.
\newblock Unsupervised discovery of interpretable directions in the {GAN}
  latent space, 2020.

\bibitem{Xia_2020}
Weihao Xia, Yujiu Yang, Jing-Hao Xue, and Wensen Feng.
\newblock Controllable continuous gaze redirection.
\newblock {\em Proceedings of the 28th ACM International Conference on
  Multimedia}, Oct 2020.

\bibitem{xia2021gan}
Weihao Xia, Yulun Zhang, Yujiu Yang, Jing-Hao Xue, Bolei Zhou, and Ming-Hsuan
  Yang.
\newblock {GAN} inversion: {A} survey, 2021.

\bibitem{xu2021generative}
Yinghao Xu, Yujun Shen, Jiapeng Zhu, Ceyuan Yang, and Bolei Zhou.
\newblock Generative hierarchical features from synthesizing images, 2021.

\bibitem{yang2021gan}
Tao Yang, Peiran Ren, Xuansong Xie, and Lei Zhang.
\newblock {GAN} prior embedded network for blind face restoration in the wild,
  2021.

\bibitem{yildirim2018disentangling}
Gökhan Yildirim, Calvin Seward, and Urs Bergmann.
\newblock Disentangling multiple conditional inputs in {GANs}, 2018.

\bibitem{zhang2018perceptual}
Richard Zhang, Phillip Isola, Alexei~A Efros, Eli Shechtman, and Oliver Wang.
\newblock The unreasonable effectiveness of deep features as a perceptual
  metric.
\newblock In {\em CVPR}, 2018.

\bibitem{zhou2019hype}
Sharon Zhou, Mitchell~L. Gordon, Ranjay Krishna, Austin Narcomey, Li Fei-Fei,
  and Michael~S. Bernstein.
\newblock {HYPE}: {A} benchmark for human eye perceptual evaluation of
  generative models, 2019.

\bibitem{zhu2020indomain}
Jiapeng Zhu, Yujun Shen, Deli Zhao, and Bolei Zhou.
\newblock In-domain {GAN} inversion for real image editing, 2020.

\bibitem{zhu2021improved}
Peihao Zhu, Rameen Abdal, Yipeng Qin, John Femiani, and Peter Wonka.
\newblock Improved {StyleGAN} embedding: {W}here are the good latents?, 2021.

\end{thebibliography}
}

\end{document}